\documentclass[dvipsnames,format=sigconf,anonymous=false,review=false]{acmart}
\AtBeginDocument{%
  \providecommand\BibTeX{{%
    Bib\TeX}}}

\usepackage{xspace}
\usepackage[linesnumbered, ruled, vlined, onelanguage]{algorithm2e}
\usepackage{amsmath,amsfonts,amsthm,mathrsfs, xfrac}
\usepackage{stmaryrd}
\usepackage{algorithmic}
\usepackage{tikz}
\usetikzlibrary{shapes, arrows, positioning}
\usepackage{float}
\usepackage{graphicx}
\usepackage{caption}
\usepackage{dirtytalk}
\usepackage{subcaption}
\usepackage{textcomp}
\usepackage{xcolor}
\def\BibTeX{{\rm B\kern-.05em{\sc i\kern-.025em b}\kern-.08em
    T\kern-.1667em\lower.7ex\hbox{E}\kern-.125emX}}

\usepackage[nameinlink, capitalize]{cleveref}
\crefname{lemma}{lemma}{lemmas}
\makeatletter
\patchcmd{\@cref}{\begingroup}{\begingroup\scshape}{}{}
\makeatother

\copyrightyear{2025}
\acmYear{2025}
\setcopyright{cc}
\setcctype{by}
\acmConference[GECCO '25]{The Genetic and Evolutionary Computation Conference 2025}{July 14--18, 2025}{Málaga, Spain}
\acmBooktitle{The Genetic and Evolutionary Computation Conference (GECCO '25),
July 14--18, 2025, Málaga, Spain}
\acmDOI{10.1145/3712256.3726380}
\acmISBN{979-8-4007-1465-8/2025/07}

\acmSubmissionID{pap368}

\newcounter{bend}
\newtheorem{lem}[bend]{Lemma}
\newtheorem*{lem*}{Lemma}

\newtheorem{thm}[bend]{Theorem}
\newtheorem*{thm*}{Theorem}
\newtheorem{rem}[bend]{Remark}

\newcommand{\LRSAO}{{\normalfont \textsc{LRSAO}}\xspace}
\newcommand{\Jump}{{\normalfont \textsc{Jump}}$_{\ell}$\xspace}
\newcommand{\Jumpnol}{{\normalfont \textsc{Jump}}\xspace}
\newcommand{\RightBridge}{{\normalfont \textsc{RightBridge}}\xspace}
\newcommand{\LeftBridge}{{\normalfont \textsc{LeftBridge}}\xspace}
\newcommand{\QLearning}{{\normalfont \textsc{Q-Learning}}\xspace}
\newcommand{\OneMax}{{\normalfont \textsc{OneMax}}\xspace}
\newcommand{\RLS}{{\normalfont \textsc{RLS}}\xspace}
\newcommand{\EARL}{{\normalfont \textsc{EA+RL}}\xspace}
\newcommand{\SOO}{{\normalfont \textsc{SOO}}\xspace}

\SetCommentSty{commentalgo}%
\SetKwComment{Comment}{/* }{ */}%
\SetKwProg{Init}{init}{}{}

\newcommand*{\N}{\mathbb{N}}
\newcommand*{\R}{\mathbb{R}}

\newcommand*{\argmax}{\mathop{\mathrm{arg\,max}}}

\newcommand*{\ens}[1]{\left\{#1\right\}}%
\newcommand*{\enstq}[2]{\left\{#1\,\middle|\,#2\right\}}%

\newcommand*{\proba}{\mathop{\mathbb{P}}}
\newcommand*{\probacond}[2]{\mathop{\mathbb{P}}\left( #1 \, \mid \, #2 \right)}
\newcommand*{\esp}{\mathop{\mathbb{E}}}
\newcommand*{\espcond}[2]{\mathop{\mathbb{E}}( #1 \, \mid \, #2 )}

\renewcommand*{\o}{\mathop{o}\limits}
\renewcommand*{\O}{\mathop{O}\limits}

\renewcommand{\theequation}{\Alph{equation}}

\begin{document}

\title{Unlearning Works Better Than You Think: \\ Local Reinforcement-Based Selection of Auxiliary Objectives}

\author{Abderrahim Bendahi}
\authornote{Equal contribution to the present paper.}
\email{abderrahim.bendahi@polytechnique.edu}
\affiliation{%
  \institution{École Polytechnique, \\
Institut Polytechnique de Paris,}
  \city{Palaiseau}
  \country{France}
}

\author{Adrien Fradin}
\authornotemark[1]
\email{adrien.fradin@polytechnique.edu}
\affiliation{%
  \institution{École Polytechnique, \\
Institut Polytechnique de Paris,}
  \city{Palaiseau}
  \country{France}
}

\author{Matthieu Lerasle}
\authornote{Senior author, supervises this research project.}
\email{matthieu.lerasle@ensae.fr}
\affiliation{%
  \institution{Institut Polytechnique de Paris,}
  \city{Palaiseau}
  \country{France}
}


\begin{abstract}
    We introduce \underline{L}ocal \underline{R}einforcement-Based \underline{S}election of \underline{A}uxiliary \underline{O}bjectives (\LRSAO), a novel approach that selects auxiliary objectives using reinforcement learning (\textsc{RL}) to support the optimization process of an evolutionary algorithm (\textsc{EA}) as in \EARL framework and furthermore incorporates the ability to unlearn previously used objectives. By modifying the reward mechanism to penalize moves that do no increase the fitness value and relying on the local auxiliary objectives, \LRSAO dynamically adapts its selection strategy to optimize performance according to the landscape and unlearn previous objectives when necessary.
    
    We analyze and evaluate \LRSAO on the black-box complexity version of the non-monotonic \Jump function, with gap parameter $\ell$, where each auxiliary objective is beneficial at specific stages of optimization. The \Jump function is hard to optimize for evolutionary-based algorithms and the best-known complexity for reinforcement-based selection on \Jump was $\O(n^2 \log(n) / \ell)$. Our approach improves over this result to achieve a complexity of $\Theta(n^2 / \ell^2 + n \log(n))$ resulting in a significant improvement, which demonstrates the efficiency and adaptability of \LRSAO, highlighting its potential to outperform traditional methods in complex optimization scenarios.

    Code is available at \href{https://github.com/FAdrien/LRSAO}{\color{cyan} https://github.com/FAdrien/LRSAO}.

\end{abstract}

\ccsdesc[500]{Computing methodologies~Sequential decision making}



\keywords{Evolutionary Algorithms, Reinforcement Learning, \EARL}


\maketitle

\section{Introduction}
        Single-objective optimization (\SOO) problems often benefit from the inclusion of auxiliary objectives alongside the primary target objective. These auxiliary objectives, mostly handcrafted~\cite{Ma1}, can enhance Random Local Search (\RLS) algorithms by helping these to traverse plateaus~\cite{Bro1} and escape or reduce the number of local optima~\cite{Kno1}. However, auxiliary objectives can sometimes be detrimental, making their dynamic selection a challenge. Traditional selection methods~\cite{And1, Ven1, Ste1} often rely on static or problem-focused approaches that lack adaptability to various optimization landscapes. Addressing this limitation, the \EARL hybrid method~\cite{Sut1, Hui1, LiP1} integrates evolutionary algorithms with reinforcement learning to dynamically select auxiliary objectives~\cite{Buz1, Buz2}. By leveraging a reinforcement learning agent to evaluate the utility of objectives, \EARL adapts the selection process based on real-time feedback, offering improved performance in monotonic optimization problems. While this hybrid method has been theoretically analyzed for monotonic functions~\cite{Ma1, Buz1, Pet1, Buz4} and successfully applied to practical scenarios~\cite{Ma1, Shi1}, its effectiveness in optimizing non-monotonic functions has not been fully explored, despite some partial progress on the \Jump function in~\cite{Ant1}, a well-studied benchmark in the literature and known to be hard to optimize due to its local optima~\cite{DoerrL24, LissovoiOW23}.
    
        In this work, we focus on single-objective optimization (\SOO) enhanced with multi-objectivization. We propose \underline{L}ocal \underline{R}einforcement-Based \underline{S}election of \underline{A}uxiliary \underline{O}bjectives (\LRSAO), a novel extension of the \EARL framework. Our approach incorporates a local-based reward mechanism which exhibits, contrary to its predecessors, an \textit{unlearning} ability. This unlearning arises from the ability of \LRSAO to discard previously useful auxiliary objectives that have become irrelevant in later stages of optimization, that is, objectives which do not bring any further improvements.
        We demonstrate the efficiency of \LRSAO on the challenging \Jump function (its black-box complexity version, see~\cite{DoerrDK14, Ant1}) where $\ell$ is the gap parameter, that is, the size of the left and right plateaus, i.e., $[0 .. \ell]$ and $[n - \ell .. n - 1]$, in which no information on the function is known, i.e., \Jump equals $0$. In this setup, each auxiliary objective offers varying benefits at different stages of the optimization process, notably in these two plateaus.
        
        Our method, leveraging a novel proof strategy for \LRSAO on plateaus, demonstrates significant improvements over the average runtime achieved in~\cite{Ant1} on \Jump (defined in subsection \hyperref[jump-def]{3.1}), reducing it from $\O(n^2 \log(n) / \ell)$ to $\Theta(n^2 / \ell^2 + n \log(n))$ without the need to restart the algorithm from scratch. These positive results on \Jump suggest the potential of \LRSAO as a robust and efficient solution for handling non-monotonic optimization problems.


    \section{Related Works}
        The use of auxiliary objectives (or helpers functions) to complement the primary objective has been a long-standing strategy in optimization research (see surveys~\cite{Ma1, SONG2024101517}). Auxiliary objectives help algorithms navigate difficult search spaces, escape local optima, and traverse plateaus. Some primary works explored static approaches, often relying on decomposing the target objective into sub-goals~\cite{10.1007/978-3-540-87700-4_4, Jensen2005} or introducing additional objectives to guide the optimization process, sometimes generated~\cite{Buz3}. These methods have been effective in certain contexts such as jobs scheduling, vertex cover or the Traveling Salesman Problem~\cite{Ma1}, but their inability to adapt to dynamically changing landscapes, where the helpfulness of auxiliary objectives can vary during optimization, led to the development of more flexible selections and designs approaches~\cite{10612125}.
    
        One notable method, \EARL, employs reinforcement learning to dynamically select auxiliary objectives based on their utility during the optimization process~\cite{Buz1, Sut1}. \EARL has been shown to exclude harmful objectives and dynamically adapt to different optimization phases, demonstrating strong theoretical and empirical performance on monotonic problems~\cite{Buz4, Buz3, Buz1}. However, its limitations in handling non-monotonic functions, such as \Jumpnol, have been identified as a key area for improvement. A notable contribution which analyzes \EARL for non-monotonic functions was made in~\cite{Ant1}. Their study focuses on optimizing the \Jump function using \EARL, considering auxiliary objectives that vary in helpfulness during different phases of optimization. The black-box \Jump function has been extensively studied in the literature of evolutionary algorithms (see~\cite{DrosteJW02, DoerrDK14, LehreW10, BuzdalovDK17}) using various approaches and is widely considered as a hard function to optimize for evolutionary-based algorithms~\cite{Ste1, DoerrL24, Bambury2024}. In~\cite{Ant1}, the authors showed that their algorithm, tailored using a restart threshold, achieves a runtime complexity of $\O(n^2 \log(n) / \ell)$, offering theoretical insights into \EARL behavior in such scenarios. However, challenges remain in improving \EARL efficiency, particularly by addressing the need for dynamic adaptation of the objectives selection over time.
    
        Our work builds on these foundations by enhancing \EARL reward mechanism. In this paper, we show that this novel mechanism \textbf{(1)} allows \LRSAO to cross plateaus at a faster rate and \textbf{(2)} avoids the need to restart from scratch the \EARL due to past mistakes thus answering two of the core limitations of~\cite{Ant1}. Besides, our algorithm \textbf{(3)} also achieves superior runtime performance on the \Jump function. This contribution to the field of evolutionary computation (\textsc{EC}) aligns with previous efforts to develop and improve reinforcement-based strategies for adaptive optimization~\cite{AntipovBD15, Buz4, Ant1}.

    \section{Problem Statement}%
        Throughout this paper, we consider bit strings $x \in \{0,1\}^n$ of length $n \geq 8$. 
        We focus on zeroth-order \emph{black-box} maximization problem of the form $x^* \in \argmax_{x \in \{0,1\}^n} f(x)$ where only \emph{partial knowledge} of the fitness value of $f$ on the current bit string $x$ and on its neighbors (the bit strings at Hamming distance $1$ of $x$) can be accessed. Following~\cite{Ant1}, the primary target $f$ is the multimodal black-box \Jump function and we use two auxiliary objectives \LeftBridge and \RightBridge which we recall in Section~\ref{sec-III-B}. These three functions are abbreviated with their first letter as \texttt{J}, \texttt{L} and \texttt{R} respectively and the global maximum of \Jump is denoted as $x^* = [1, \ldots, 1]$.

 \subsection{The Auxiliary and Target Objectives} \label{sec-III-B}
                Given a positive integer $n \geq 8$, we work on the hypercube $\{0,1\}^n$ of the bit strings of length $n$ and all three objectives defined below are from $\ens{0, 1}^n \to [ 0 .. n ]$. We write $x = (x_1, \ldots, x_n)$ for some bit string $x \in \ens{0, 1}^n$. Below, we recall the unimodal \OneMax function, \Jump, and also \LeftBridge and \RightBridge as defined in~\cite{Ant1}.
                    \begin{align*}
                        \OneMax \colon & x \mapsto \| x \|_1 = \sum_{i = 1}^n x_i, \\
                        \Jumpnol_{\ell} \colon & x \mapsto \begin{cases} \| x \|_1, & \text{if $\| x \|_1 \in [ \ell + 1 .. n - \ell - 1 ] \cup \{ n \}$;} \\ 0, & \text{otherwise.} \end{cases} \tag*{(\texttt{J})}\label{jump-def}
                    \end{align*}
                
                The two auxiliary objectives, adapted from~\cite{Ant1}, are
                    \begin{align*}
                        \LeftBridge & \colon x \mapsto \begin{cases} \| x \|_1, & \text{if $\| x \|_1 \in [ 0 .. \ell + 1 ]$;} \\ 0, & \text{otherwise;} \end{cases} \\
                        \RightBridge & \colon x \mapsto \begin{cases} \| x \|_1, & \text{if $\| x \|_1 \in [ n - \ell - 1 .. n ]$;} \\ 0, & \text{otherwise;} \end{cases}
                    \end{align*}
                 where $\ell \in \left[ 2 \, .. \!\left\lfloor \frac{n - 1}{2} \right\rfloor - 2 \right]$ is a parameter controlling the size of the left and right plateaus of \Jump (notably, $\ell < \frac{n}{2} - 1$). The case $\ell = 1$ does not adequately highlight the learning process of \LeftBridge on the left plateau and its unlearning on the right plateau (in order to learn \RightBridge instead). In this case the right plateau becomes too narrow (of size $1$) and crossing it can be done with an average time of $\O\left(n^2\right)$ independently of which objective is used. Also, when $\ell = \left\lfloor \frac{n - 1}{2} \right\rfloor - 1$, \Jump is reduced to three isolated points at $\ell + 1$, $n - \ell - 1$ and $n$ and became useless for the \textsc{Q-Learning} agent.

                While the \Jump function is the same as in~\cite{Ant1}, the two auxiliary objectives have been slightly adapted to include the endpoints at $\ell + 1$ and $n - \ell + 1$ for \LeftBridge and \RightBridge respectively.
                
\begin{figure}
                    
                    \begin{subfigure}{.45\linewidth}
                        \centering
                        \includegraphics[width=4cm]{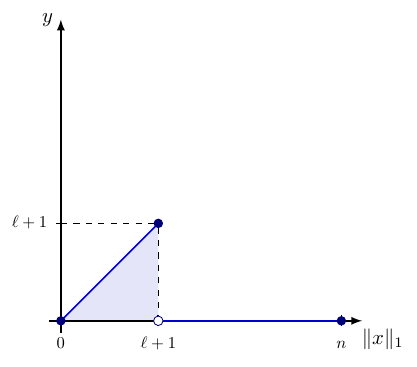}
                        \caption{\footnotesize\textit{The {\normalfont \LeftBridge} objective}}
                        \label{fig:sub1}
                        \end{subfigure}
                    \hfill
                    \begin{subfigure}{.45\linewidth}
                        \centering
                        \includegraphics[width=4cm]{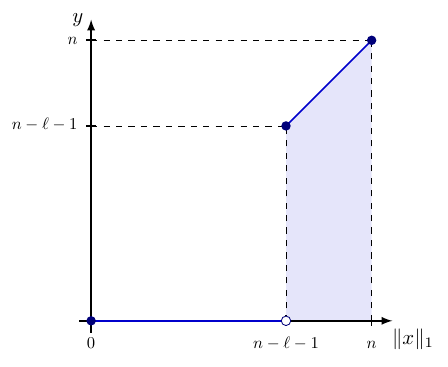}
                        \caption{\footnotesize\textit{The {\normalfont \RightBridge} objective}}
                        \label{fig:sub2}
                    \end{subfigure}\\[1ex]
                    \begin{subfigure}{\linewidth}
                        \centering
                        \includegraphics[width=4cm]{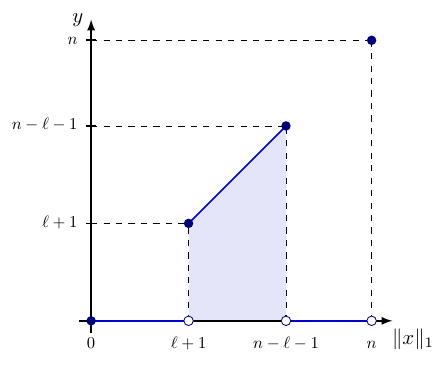}
                        \caption{\textit{The {\normalfont \Jump} objective}}
                        \label{fig:sub3}
                    \end{subfigure}
                    \caption{\footnotesize\small\textit{The three objectives.}}
                    \label{fig:objectives}
                \end{figure}

                These functions are plotted in~\cref{fig:objectives}.
                
        \subsection{The \LRSAO Algorithm}
            The \LRSAO algorithm in~\cref{alg:alg1} is based on the \EARL algorithm from~\cite{Ant1} with some changes in the reward mechanism. As this algorithm incorporates a reinforcement learning \textit{agent} to select an objective, we introduce $\alpha \in (0, 1)$ as the learning rate of the \textsc{Q-Learning} agent and $\gamma \in (0, 1)$ as the discount factor used to update the $Q$-table. The state space in which the \textit{agent} evolves is denoted by $\mathcal{S}$ (for \Jump we have $\mathcal{S} = \ens{0} \cup [ \ell + 1 .. n - \ell - 1 ] \cup \ens{n}$).
            
            During iteration $t$, the current bit string $x_t$ undergoes a mutation, which consists in flipping one of its bits with uniform probability, and gives bit string $x_{\texttt{new}}$. Then, the auxiliary objective $f_t \in \ens{\texttt{L}, \texttt{J}, \texttt{R}}$ which maximizes the entry $Q_t[s_t, \cdot]$ is selected (in case of a tie, $f_t$ is chosen uniformly among the actions maximizing the entry $Q_t[s_t, \cdot]$). Then, based on the fitness value of $x_t$ and $x_{\texttt{new}}$, one of them is chosen as the new individual $x_{t + 1}$ and a reward $r_{t + 1}$ is assigned. The key difference with \EARL is how we define $r_{t + 1}$:\hypertarget{reward-def}{}
                \[ r_{t + 1} = \begin{cases}
                    0, & \text{if $f_t(x_{\texttt{new}}) < f_t(x_t)$;} \\
                    -r, & \text{if $f_t(x_{\texttt{new}}) = f_t(x_t)$;} \tag*{(R)}\label{reward-def}\\
                    f_t(x_{\texttt{new}}) - f_t(x_t), & \text{otherwise;}
                \end{cases} \]
            where $r > 0$ is a penalty for having the same fitness value. As the offspring $x_{\texttt{new}}$ \emph{always} has a different position than its parent $x_t$, that is $\| x_t \|_1 \neq \| x_{\texttt{new}} \|_1$, we may then see $r$ as a penalty for moving on a plateau of $f_t$. This penalty allows \LRSAO to quickly discard objectives when necessary and is in the core of its \textit{unlearning} ability.

            \begin{algorithm}
                \captionof{algocf}{\textit{\LRSAO, combining \EARL with a local target reward and plateau penalty.}}
                \label{alg:alg1}
                \DontPrintSemicolon           
                \SetKwProg{Init}{Initialization}{:}{}
                \Init{}{
                        $t \gets 0$\;
                        $x_0 \gets [0, \ldots, 0]$, a $1 \times n$ vector, filled with zeros\;
                        $s_0 \gets \Jumpnol_{\ell}(x_0)$\;
                        $Q_0 \gets (n + 1) \times 3$ matrix, filled with zeros\;
                }

                \vspace{\baselineskip}

                \While{$s_t < n$}{
                        $x_{\texttt{new}} \gets \textsc{RandomOneBitFlip}(x_t)$\;
                        $f_t \gets \argmax_{a \in \mathscr{A}} Q_t[s_t, a]$ \tcp*[l]{Break tie if needed.}

                        \eIf{$f_t(x_{\textup{\texttt{new}}}) \geq f_t(x_t)$}{
                                $r_{t + 1} \gets f_t(x_{\texttt{new}}) - f_t(x_t)$\;
                                $x_{t + 1} \gets x_{\texttt{new}}$\;

                                \tcp*[h]{Penalty reward for plateaus.}\;
                                \If{$r_{t + 1} = 0$}{
                                        $r_{t + 1} \gets -r$;
                                }
                        }{
                                \tcp*[h]{The move to $x_{\texttt{new}}$ is rejected.}\;
                                $x_{t + 1} \gets x_t$\;
                                $r_{t + 1} \gets 0$\;
                        }

                        $s_{t + 1} \gets \Jumpnol_{\ell}(x_{t + 1})$\;

                        \tcp*[h]{Update the $Q$-table by first duplicate $Q_t$ to form $Q_t + 1$.}\;
                        $Q_{t + 1}[s_t, f_t] \gets (1 - \alpha) Q_t[s_t, f_t] + \alpha(r_{t + 1} + \gamma \cdot \max_{a \in \mathscr{A}} Q_t[s_{t + 1}, a])$\;

                        $t \gets t + 1$;
                }
            \end{algorithm}
                    
    \section{Notation and the Main Assumption}
        \subsection{Notation}
        Some notations introduced in~\cref{alg:alg1} and collected in~\cref{tab:notation} are defined here along with other symbols (stopping times and events) to ease the runtime analysis of \LRSAO.
        
        In the present work, we denote by $x_t$, $s_t$ and $f_t$ the bit string, the state\footnote{The word \textit{state} refers to elements of the state space $\mathcal{S}$ and should not be confused with the \textit{position} or \emph{individual}, corresponding to $\| x_t \|_1 \in [ 0 .. n ]$.} and the action taken at time $t$. By definition, $s_t = \Jumpnol_{\ell}(x_t)$ and the reward received by the agent at time $t$ is noted $r_{t + 1}$. Here, $Q_t$ is the $Q$-table used during iteration $t$ (to choose $f_t$ for instance).

        On the other hand, in order to express some events and conditioning, we introduce additional notations. Given $t \geq 0$ and $p \in [ 0 .. n ]$, we let $\texttt{H}^p_t$ be the event \say{\emph{hitting\footnote{That is, $\| x_t \|_1 \neq p$ while $\| x_{t + 1} \|_1 = p$, i.e., position $p$ is hit at the end of iteration $t$.} position $p$ at time $t$}}, while the other events are related to actions taken at time $t$, namely $\texttt{L}^+_t$ (resp. $\texttt{J}^+_t$ and $\texttt{R}^+_t$ for \Jump and \RightBridge) the event \say{\emph{choosing \LeftBridge at time $t$, i.e., $f_t = \texttt{L}$, with $\| x_{\texttt{new}} \|_1 > \| x_t \|_1$}} and $\texttt{L}^-_t$ (resp. $\texttt{J}^-_t$ and $\texttt{R}^-_t$) the event \say{\emph{choosing \LeftBridge at time $t$, i.e., $f_t = \texttt{L}$ with $\| x_{\texttt{new}} \|_1 < \| x_t \|_1$}}. The sign $\pm$ on these events means that the proposed mutation is directed toward $x^*$ ($+$) or toward $0^n$ ($-$). Also, for an objective $f \in \mathscr{A}$, we write $f^+_{t, \texttt{plateau}}$ to denote the occurrence of the event $f^+_t$ at some time $t \geq 0$ in a plateau\footnote{This means, being in a plateau of \Jump and the event $f^+_t$ occurs.} of \Jump.
        
        Besides these notations, we introduce some stopping times to split a run of~\cref{alg:alg1} into three phases. The total runtime is $T = \inf \enstq{t \geq 0}{x_t = x^*} \in \N_0 \cup \ens{+\infty}$ (first hitting time of $x^*$) and is split into times: $T_1$ the first hitting time of state $\ell + 1$, i.e., the first time we leave the left plateau of \Jump, $T_2$ the time from the end of the first phase until we first reach the right plateau of \Jump (i.e., $\| x_t \|_1 = n - \ell$) and $T_3$ the remaining time until $x^*$ is found. Throughout this study, we are interested in estimating an upper bound on $\esp(T)$, the average \textit{total} runtime of \LRSAO. To do so, we use the fact that $T = T_1 + T_2 + T_3$ and we upper bound, in subsections~\ref{par-D}, \ref{par-E} and~\ref{par-F}, the quantities $\esp(T_1)$, $\esp(T_2)$ and $\esp(T_3)$, the average runtime of \LRSAO on the first, second and third phase.

        \begin{table}
          \caption{Summary of notation}
          \label{tab:notation}
          \begin{tabular}{cl}
            \toprule
            Time & Meaning \\
            \midrule
            $T$ & The overall runtime \\
            $T_1$, $T_2$, $T_3$ & Runtime of the first, second and third phase \\
            \midrule
            Domain & Definition \\
            \midrule 
            $\mathscr{A}$ & The action space where $\mathscr{A} = \{\texttt{L}, \texttt{J}, \texttt{R}\}$ \\ 
            $\mathcal{S}$ & The state space, here $\mathcal{S} = \ens{0} \cup [ \ell + 1 .. n - \ell - 1 ] \cup \ens{n}$ \\ 
            \midrule
            Symbol & Meaning \\
            \midrule
            \texttt{L}, \texttt{J}, \texttt{R} & The actions \LeftBridge, \Jump and \RightBridge \\
            $Q_t$ & The $Q$-table at time $t$ \\
            $x^*$ & The global maximum of \Jump, $x^* = [1, \ldots, 1]$ \\
            $x_t$ & The bit string at time $t$, $x_t \in \{0, 1\}^n$ \\
            $x_{\texttt{new}}$ & The bit string $x_t$ with one of its bits flipped (mutation) \\
            $s_t$ & State at time $t$ with $s_t = \Jumpnol_{\ell}(x_t)$ \\ 
            $f_t$ & Action taken at time $t$ and $f_t \in \mathscr{A}$ \\
            $r_{t + 1}$ & Reward at time $t$, defined in~\ref{reward-def}\\
            $r$ & Reward penalty when crossing a plateau ($r > 0$) \\
            $\alpha$ & Learning rate of the \QLearning agent, $\alpha \in (0, 1)$ \\ 
            $\gamma$ & Discount factor, $\gamma \in (0, 1)$ \\
            $\mathcal{H}_n$ & The $n$-th harmonic number, $\mathcal{H}_n = \sum_{k = 1}^n \frac{1}{k}$ \\ 
            \midrule
            Event & Definition \\
            \midrule
            $\texttt{a}^+_t$ & Choose $a \in \mathscr{A}$ at time $t$ and move toward $x^*$ \\
            $\texttt{a}^-_t$ &  Choose $a \in \mathscr{A}$ at time $t$ and move away from $x^*$ \\ 
            $\texttt{a}^+_{t, \texttt{plateau}}$ & The event $a^+_t$ occurs in a plateau of \Jump at time $t$ \\
            $\texttt{H}_t^p$ & Hit position $p \in [0..n]$ at time $t$ \\
          \bottomrule
        \end{tabular}
        \end{table}

        \subsection{The Main Assumption}
            In what follows, the penalty $r > 0$ satisfies
                \begin{equation} 
                    \left( \frac{1}{\alpha (1 - \gamma)} - 1 \right) (n - \ell - 1) < r < \frac{1}{\alpha \gamma},\tag*{(H)}\label{penalty-hyp}
                \end{equation}
            given $0 < \alpha, \gamma < 1$ such that $\frac{1 - \gamma}{\gamma (1 - \alpha (1 - \gamma))} > n - \ell - 1$.
            
            Notice that it is enough to take $0 < \gamma \leq \frac{1}{n + 1}$ without any assumptions on $\alpha$ or even $\ell$. The lower bound on the penalty $r$ in~\ref{penalty-hyp} ensures that \LRSAO can quickly discard objectives that turn out to be non-relevant in the current region of optimization while the upper bound guarantees that \LRSAO cannot be stuck in state $n - \ell - 1$ in case $Q[0, \texttt{L}], Q[0, \texttt{J}]$ and $Q[0, \texttt{R}]$ are negative. This addresses one of the issues of \EARL~\cite{Ant1}.
            
            The inequalities~\ref{penalty-hyp} are crucial in our study and suggest a greedy behavior of the \QLearning agent, i.e., maximizing the gain from auxiliary objectives in short time horizon. This outlines a general idea the authors wanted to convey through this work: \textit{let the auxiliary objectives guide you through the landscape of the target objective}.
        
            \section{Runtime Analysis of \LRSAO}
                \subsection{Main Result}

            Our main result is the~\cref{thm:4} below. It provides an upper bound on the average runtime of \LRSAO (see~\cref{alg:alg1}) on \Jump.
            \begin{thm}[Total Average Runtime] \label{thm:4}
                The \LRSAO algorithm optimizes the black-box \Jump function with an average runtime of
                    \[ \esp(T) = \Theta\left( \frac{n^2}{\ell^2} + n \log(n) \right). \]
            \end{thm}

            The runtime of \LRSAO as presented in~\cref{thm:4} supersedes the previous $\O(n ^2 \log(n) / \ell)$ average runtime of \EARL~\cite{Ant1}. Moreover, \LRSAO does not need any restart mechanism whose cutoff time might be benchmark-specific and might require manual tuning.

            In subsections~\ref{sec:gen-results}, \ref{par-D}, \ref{par-E} and~\ref{par-F}, our goal is to prove the above theorem by computing an upper bound on the expectation of each time $T_1$, $T_2$ and $T_3$ separately. We distinguish two cases according to the landscape of the \Jump function: \textbf{(1)} the two plateaus (left and right) and \textbf{(2)} the middle slope. For the increasing slope of \Jump (phase $2$), we follow the same approach as in~\cite{Ant1} by showing in~\cref{lemma:Learning lemma} that~\cref{alg:alg1} cannot visits each state $s \in [ \ell + 3 .. n - \ell - 2 ]$ more than five times which is enough to upper bound $\esp(T_2)$. For the plateaus, we introduce a novel strategy which differs from the one from~\cite{Ant1} relying on the multiplicative drift theorem~\cite{DoerrJW12algo}. Instead, in subsection~\ref{subsec:lem-plateau}, we prove~\cref{lem-plateau}, a key lemma in our approach and based on that, we then split the total time on a plateau in two. One is the time needed for the \textsc{RL} agent to learn the best objective $f$ to use in the plateau (exploration phase) while the other is the remaining time (exploitation phase). During the exploitation phase, we show that the \textsc{RL} agent constantly used this best objective $f$ until it leaves the region. For the lower bound, as \LRSAO relies on the \textsc{RandomOneBitFlip} operator to produce a mutation then it needs $\Omega(n \log(n))$ calls to optimizes \Jump. We show in subsection~\ref{par-F} that $\esp(T) = \Omega(n^2 / \ell^2)$ and combining both lower bounds leads to $\esp(T) = \Omega(\max\{n^2 / \ell^2, n \log(n)\}) = \Omega(n^2 / \ell^2 + n \log(n))$.

    
        \subsection{A Global Upper Bound and Some Properties}\label{sec:gen-results}
           First, we provide an upper bound on the entries of the $Q$-table to show that these entries do not blow up to $+\infty$ over time.
            \begin{lem}\label{lemma:inequalities}
                Let $t \in [ 0 .. T - 1 ]$, $s \in \ens{0} \cup [ \ell + 1 .. n - \ell - 1 ]$ and $a \in \mathscr{A} = \ens{{\normalfont \texttt{L}}, {\normalfont \texttt{J}}, {\normalfont \texttt{R}}}$ then
                    \[ Q_t[s, a] < \frac{n - \ell - 1}{1 - \gamma}. \]
            \end{lem}
    
            \begin{proof}
               By induction on $t$, all entries of the $Q$-table are zeros at $t = 0$, which is less than $\frac{n - \ell - 1}{1 - \gamma}$. Now, if at iteration $t < T - 1$ the inequality is satisfied for all values of $s$ and $a$ then, as $t + 1 < T$, we have $s_t \neq n$ and $s_{t + 1} \neq n$ so $r_{t + 1} \leq n - \ell - 1$ (the highest achievable reward unless $n$ is reached) and when the $Q$-table is updated,
                \begin{align*}
                   Q_{t + 1}[s_t, f_t] & = (1 - \alpha) Q_t[s_t, f_t] + \alpha(r_{t + 1} + \gamma \max_{a \in \mathscr{A}} Q_t[s_{t + 1}, a]) \\ 
                    & < (1 - \alpha) \frac{n - \ell - 1}{1 - \gamma} + \alpha \left( r_{t + 1} + \gamma \frac{n - \ell - 1}{1 - \gamma} \right) \\ 
                    & \leq (1 - \alpha) \frac{n - \ell - 1}{1 - \gamma} + \alpha (n - \ell - 1) + \alpha \gamma \frac{n - \ell - 1}{1 - \gamma} \\ 
                    & = \frac{n - \ell - 1}{1 - \gamma},
                \end{align*}
                as desired. The other entries of the $Q$-table are unchanged.
            \end{proof}

            \begin{lem}[$Q$-Table and a Local Maximum]\label{lem:local-maximum}
                For any set $\mathscr{A}$ of objectives, if state $s \in \mathcal{S}$ is a strict local maximum of an objective $a \in \mathscr{A}$ then, for any time $t \geq 0$, $Q_t[s, a] = 0$.
            \end{lem}

            \begin{proof} (Sketch) 
                Initially all entries of the $Q$-table are set to zero and as $s$ is a strict local maximum of $a$, every offspring $x_{\texttt{new}}$ will be rejected if $a \in \mathscr{A}$ is chosen. Based on this remark, we then proceed by induction on $t$. A full proof can be found in the appendix.
            \end{proof}

            \subsection{Key Lemma for the Plateaus}\label{subsec:lem-plateau}
            \begin{lem} \label{lem-one-positive-per-state}
                Let $\mathscr{A}$ be a set of objectives then, for any state $s \in \mathcal{S}$ and any time $t \geq 0$ there exists at most one $a \in \mathscr{A}$ such that $Q_t[s, a] > 0$.
            \end{lem}
    
            \begin{proof}
                Recall that $Q_0$ is set to zero initially. Now, assume there exists $t_1 > 0$, a state $s$ and actions $a_0, a_1 \in \mathscr{A}$, with $a_0 \neq a_1$, such that $Q_{t_1}[s, a_0] > 0$ and $Q_{t_1}[s, a_1] > 0$. Without loss of generality, suppose $t_1$ is minimal, i.e., for any $0 \leq t < t_1$ and any state $s$, at most one entry of $Q_t[s, \cdot]$ is positive. Since in~\cref{alg:alg1}, exactly one entry of the $Q$-table is updated each iteration, we can assume entry $[s, a_1]$ to be the one updated during iteration $t_1 - 1$, hence $s_{t_1 - 1} = s$ and $f_{t_1 - 1} = a_1$. Moreover, by minimality of $t_1$ we should have $Q_{t_1 - 1}[s, a_0] > 0$ and $Q_{t_1 - 1}[s, a_1] \leq 0$, contradicting the fact that objective $a_1$ has been selected during iteration $t_1 - 1$.
            \end{proof}
    
            For non-negative integers $a < b$, we say that $[a.. b]$ is a \emph{plateau} of some objective $f \in \mathscr{A}$ if $f$ is a constant across all positions $p \in [a .. b]$, that is, $f$ is a constant over all bit strings $x \in \{0, 1\}^n$ such that $\| x \|_1 \in [a .. b]$. Moreover, an objective $f$ is said to be \emph{strictly increasing} over $[a .. b]$ if for any two bit strings $x$ and $y$ such that $a \leq \| x \|_1 < \| y \|_1 \leq b$ we have $f(x) < f(y)$.
            
            Our strategy to upper bound the average runtime of \LRSAO to cross a plateau $[a.. b]$ consists of splitting the crossing phase in two sub-phases, based on the occurrence of a certain event $E_t$ at time $t \geq 0$. The next lemma makes clear what $E_t$ could be.
            \begin{lem} \label{lem-plateau}
                Let $\mathscr{A}$ be a set of objectives from $\ens{0, 1}^n \to \R$ with target function $\mathscr{F} \in \mathscr{A}$ such that $[a.. b]$, with $a < b < n$ non-negative integers, is a plateau of $\mathscr{F}$ and $\mathscr{F}$ is a constant equal to $c \in \R$ over $[a .. b]$. Assume there exists an objective $f \in \mathscr{A}$ strictly increasing over the plateau $[a.. b]$, then
                \begin{enumerate}
                    \item for any time $t \geq t_s$ of a walk $\mathcal{W}$ over $[a.. b]$,
                        \[ Q_t[c, f] \geq (1 - \alpha (1 - \gamma))^{\ell(t)} Q_{t_s}[c, f], \tag*{(I)}\label{lem:plateau-ineq} \]
                        where $t_s$ is the starting time of the walk (such that $s_{t_s} = c$) and $\ell(t)$ is the number of times the objective $f$ is used between iterations $t_s$ and $t - 1$ (inclusive).
                    
                    \item if the event $f^+_{t, \normalfont \texttt{plateau}}$ occurred at some time $t = t_0 \geq 0$ and $\| x_{t_0 + 1} \|_1 = k \in (a.. b]$ then $f$ is always selected until we leave the plateau $[a .. b]$ to reach $b + 1$ and the expected time $T_{b + 1}$ to leave $[a .. b]$ from position $k$ is
                    \[ \esp(T_{b + 1}) = n \left( \mathcal{H}_{n - k} - \mathcal{H}_{n - (b + 1)} \right), \]
                where $\mathcal{H}_n = \sum \limits_{k = 1}^n \frac{1}{k}$ is the $n$-th harmonic number.
                \end{enumerate}
            \end{lem}
    
            \begin{proof}
                For the first statement, we proceed by induction. Let $\mathcal{W}$ be a walk starting at time $t_s \geq 0$ on the plateau $[a.. b]$ of the target objective $\mathscr{F}$. As $\ell(t_s) = 0$ then inequality~\ref{lem:plateau-ineq} holds at time $t = t_s$. Now assume inequality~\ref{lem:plateau-ineq} holds along the walk $\mathcal{W}$ up to time $t < t_e - 1$ where $t_e$ is the ending time of $\mathcal{W}$ on $[a.. b]$, that is, the first time $t \geq t_s$ such that $\| x_t\|_1 \notin [a .. b]$. As $t_s < t + 1 < t_e$ then $\| x_t \|_1, \| x_{t + 1}  \|_1 \in [a .. b]$ so $s_t = c = s_{t + 1}$. Now, either $f_t \neq f$ in which case $\ell(t + 1) = \ell(t)$ and $Q_{t + 1}[c, f] = Q_t[c, f]$ so~\ref{lem:plateau-ineq} holds. Otherwise, if $f_t = f$, that is, $Q_t[c, f] = \max_{f' \in \mathscr{A}} Q_t[c, f']$ and as objective $f$ is \textit{strictly increasing} over $[a.. b]$ then $r_{t + 1} \geq 0$ hence
                    \begin{align*} Q_{t + 1}[s_{t + 1}, f_t] & = Q_{t + 1}[c, f] \\ 
                    & = (1 - \alpha) Q_t[c, f] + \alpha \left( r_{t + 1} + \gamma Q_t[c, f] \right) \\
                    & \geq (1 - \alpha (1 - \gamma)) Q_t[c, f] \\ 
                    & \geq  (1 - \alpha (1 - \gamma))^{\ell(t) + 1} Q_{t_s}[c, f],
                    \end{align*}
                since $s_{t + 1} = c = s_t$ and $\ell(t + 1) = \ell(t) + 1$. Thus inequality~\ref{lem:plateau-ineq} holds and the first statement follows by induction over the walk $\mathcal{W}$.
                
                For the other statement, let $T_{\mathrm{end}} = (t_0 + 1) + T_{b + 1}$ be the first time when we leave $[a .. b]$. We show by induction on $t \in [t_0 + 1.. T_{\mathrm{end}} - 1]$ that $(H_t):$ \say{\emph{$s_t = s$ and $Q_t[c, f] > \max_{f' \in \mathscr{A} \setminus \ens{f}} Q_t[c, f']$}} holds. As $f^+_{t, \normalfont \texttt{plateau}}$ occurred at iteration $t = t_0$ and $s_{t_0} = c = s_{t_0 + 1}$ then,
                    \begin{align*}
                        Q_{t_0 + 1}[s_{t_0}, f_{t_0}] & = Q_{t_0 + 1}[c, f] \\
                        & = (1 - \alpha) Q_{t_0}[c, f] + \alpha \left( r_{t_0 + 1} + \gamma Q_{t_0}[c, f] \right) \\ 
                        & = (1 - \alpha (1 - \gamma)) Q_{t_0}[c, f] + \alpha r_{t_0 + 1}.
                    \end{align*}
                Then, as $f$ is \textit{strictly increasing} over $[a.. b]$ and $f(x_{\texttt{new}}) > f(x_{t_0})$, we have $r_{t_0 + 1} > 0$. Now, if $Q_{t_0}[c, f] \leq 0$ then as $0 < 1 - \alpha (1 - \gamma) < 1$ we obtain $Q_{t_0}[c, f] \leq (1 - \alpha (1 - \gamma)) Q_{t_0}[c, f]$ so 
                    \[ Q_{t_0 + 1}[c, f] > Q_{t_0}[c, f] \geq  \max_{f' \in \mathscr{A}} Q_{t_0}[c, f']. \]
                Otherwise, if $Q_{t_0}[c, f] > 0$ then $Q_{t_0 + 1}[c, f] > 0$ and by~\cref{lem-one-positive-per-state}, $0 \geq \max_{f' \in \mathscr{A} \setminus \ens{f}} Q_{t_0 + 1}[c, f']$. This proves the base case and shows that $f$ is chosen at time $t_0 + 1$. Now, if at time $t_1 < T_{\mathrm{end}} - 1$ hypothesis $(H_{t_1})$ holds then $s_{t_1} = c$ and $f_{t_1} = f$ is selected during iteration $t_1$. Either $f^-_{t_1, \texttt{plateau}}$ occurs but as $f$ is \textit{strictly increasing} over $[a.. b]$, the move to $x_{\texttt{new}}$ is rejected so $s_{t_1 + 1} = c$, $r_{t_1 + 1} = 0$ and
                \[ Q_{t_1 + 1}[c, f] = (1 - \alpha (1 - \gamma)) Q_{t_1}[c, f] > \max_{f' \in \mathscr{A} \setminus \{ f\}} Q_{t_1 + 1}[c, f'], \]
                since either $Q_{t_1}[c, f] > 0$ hence $Q_{t_1 + 1}[c, f] > 0$ and we are done by~\cref{lem-one-positive-per-state}, or $0 \geq Q_{t_1}[c, f]$ but then $Q_{t_1 + 1}[c, f] \geq Q_{t_1}[c, f]$ and $(H_{t_1+1})$ follows. Otherwise, if $f^+_{t_1, \texttt{plateau}}$ occurs, as $t_1 + 1 < T_{\mathrm{end}}$ we still have $s_{t_1 + 1} = c$ and, exactly as we did in the base case, we obtain the desired inequality $Q_{t_1 + 1}[c, f] > \max_{f' \in \mathscr{A} \setminus \ens{f}} Q_{t_1 + 1}[c, f']$.
    
                We have shown that for any time $t \in [t_0 + 1.. T_{\mathrm{end}} - 1]$ we stay on the plateau $[a.. b]$ and we always chose objective $f$. Since $f$ is \textit{strictly increasing} over $[a.. b]$, we cannot go backward hence, at each iteration, either we stay on the current position $m = \| x_t \|_1$ or we move to position $m + 1$. This gives the transition probabilities shown in~\cref{fig:trans-probs-1}.
                        \begin{figure}
                            \centering
                            \resizebox{4cm}{!}{%
                                \begin{tikzpicture}[->, >=stealth']
                                    \node[circle, minimum size = 1.2cm, draw](0) at (0, 0) {$m$};
                                    \node[circle, minimum size = 1.2cm, draw](1) at (3, 0) {$m + 1$};
                                    \path (0) edge [loop below] node [below] {$\frac{m}{n}$} (0);
                                    \path (0) edge [bend left] node [above] {$\frac{n - m}{n}$} (1);
                                \end{tikzpicture}
                            }%
                            \caption{\small\textit{Transitions probabilities between positions $m$ and $m + 1$.}}
                            \label{fig:trans-probs-1}
                            \vspace*{-0.4cm}
                        \end{figure}
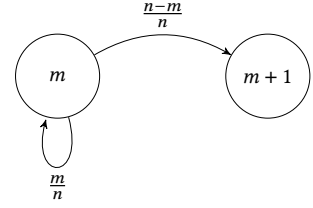
                
                If we let $T^+_m$ to be the time needed to go from $m$ to $m + 1$ then, as we cannot go backward, we obtain
                    \[ T_b = \sum_{m = k}^b T_m^+, \]
                and, as we always take objective $a$, every $T^+_m$ is the first success in i.i.d. Bernoulli trials of parameter $p = \frac{n - m}{n}$, so $\esp(T^+_m) = \frac{n}{n - m}$ thus
                    \[ \esp(T_b) = \sum_{m = k}^b \esp(T^+_m) = n \cdot \sum_{m = n - b}^{n - k} \frac{1}{m} = n \left( \mathcal{H}_{n - k} - \mathcal{H}_{n - (b + 1)} \right), \]
                as desired.
            \end{proof}
    
        \subsection{The First Phase: Learning \LeftBridge} \label{par-D}
    
            Initially, we start at $x_0 = [0, \ldots, 0]$ so in the first plateau of \Jump and all entries of the $Q$-table are set to zero. Consider the event
                \[ E_t^1 = \texttt{H}_t^{\ell + 1} \cup \texttt{L}^+_{t, \texttt{plateau}}, \]
            namely \say{\textit{use {\normalfont $\texttt{L}^+$} in the plateau or hit $\ell + 1$, at time $t$}}. Let $T_1^1$ be the first time $t$ where $E_t^1$ occurs (it occurs almost surely for some finite time $t \geq 0$), and $T_1^2$ the remaining time until the end of the first phase so that $T_1 = T_1^1 + T_1^2$. The next lemma helps to bound $\esp(T_1^1)$.
            \begin{lem}[Few Mistakes Lemma] \label{lem:NotMuchMistakes}
                There exists at most one $t_{\normalfont \texttt{J}}$ and at most one $t_{\normalfont \texttt{R}}$ in $[ 0 .. T_1 - 2 ]$, such that
                   \[ f_{t_{\normalfont \texttt{J}}} = {\normalfont \texttt{J}} \text{ and } f_{t_{\normalfont \texttt{R}}} = {\normalfont \texttt{R}}, \]
                and for any $0 \leq t \leq T_1$, both $Q_t[0, \texttt{J}]$ and $Q_t[0, \texttt{R}]$ lie in $\{0, -\alpha r\}$.
                
                Moreover $f_{T_1 - 1} = {\normalfont \texttt{L}}$ whenever $T_1$ is finite.
            \end{lem}
            \begin{proof}
                (Sketch) By~\Cref{lem-plateau} $(1)$, for any time $0 \leq t < T_1$, as we stay in the plateau $[0 .. \ell]$, we have $Q_t[0, \texttt{L}] \geq 0$ and by inequalities~\ref{penalty-hyp}, if one chooses $a \in \{\texttt{J}, \texttt{R}\}$ at time $0 \leq t < T_1 - 1$ then $Q_{t + 1}[0,a] < 0$. We conclude using $\ell \geq 2$, i.e., that at least $3$ steps are needed to leave the plateau $[0..\ell]$. A detailed proof can be found in the appendix.
            \end{proof}

            We can now state our main result.
            \begin{thm}[Runtime of the First Phase]\label{thm:phase1}
                We have:
                \begin{align*}
                    \esp(T_1) & = n \ln \left( \frac{1}{1 - \frac{\ell + 1}{n}} \right) + \frac{1}{2} - \frac{1}{2 \left( 1 - \frac{\ell + 1}{n}  \right)} + \o_{n \to +\infty}(1) \\ 
                    & \leq 2 (\ell + 1) \ln(2).
                \end{align*}
            \end{thm}
        
            \begin{proof} (Sketch)
                The first iteration results into one of two scenarios, either $\texttt{L}^+_0$ occurs (and by~\cref{lem-plateau}, \texttt{L} is selected until the end of the phase) or, $\texttt{J}^+_0 \cup \texttt{R}^+_0$ occurs, say it is $x^+_0$ where $x \in \ens{\texttt{J}, \texttt{R}}$ and let $y \in \ens{\texttt{J}, \texttt{R}} \setminus \ens{x}$ be the other objective. At time $t = 1$ we are in position $1$ and in the second scenario $x$ cannot be selected anymore according to~\cref{lem:NotMuchMistakes}. This leads to the transition probabilities shown in~\cref{fig-transition-probs}
                        \begin{figure}
                            \centering
                            \resizebox{6cm}{!}{%
                                \begin{tikzpicture}[->, >=stealth', auto, node distance = 4cm]
                                    \node[circle, minimum size = 1.2cm, draw](0) at (0, 0) {$0$};
                                    \node[circle, minimum size = 1.2cm, draw](1) at (3, 0) {$1$};
                                    \node[circle, minimum size = 1.2cm, draw](2) at (6, 0) {$2$};
                
                                    \path (1) edge [loop below] node [below] {$\frac{1}{2 n}\ (\texttt{L}^-)$} (1);
                                    \path (1) edge [bend right] node [above] {$\frac{1}{2 n}\ (y^-)$} (0);
                                    \path (1) edge [bend left] node [above] {$\frac{2 (n - 1)}{2n}\ (\texttt{L}^+ \cup y^+)$} (2);
                                \end{tikzpicture}
                            }%
                            \caption{\small\textit{Transitions probabilities between $0$, $1$ and $2$ at $t = 1$.}}
                            \label{fig-transition-probs}
                            \vspace*{-0.4cm}
                        \end{figure}
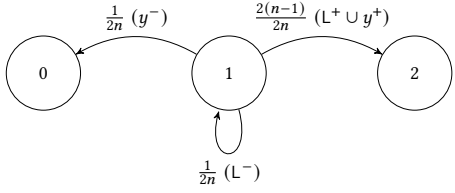
                    where we remove the \textit{time index} on the events $y^\pm$ and $\texttt{L}^\pm$. Let $T^{\pm}_1 \in \N_0 \cup \ens{+\infty}$ the time taken to leave $1$, then
                        \[ \esp(T^{\pm}_1) = \frac{1}{1 - \frac{1}{2 n}} = \frac{2 n}{2 n - 1}, \]
                    which is finite thus $T^{\pm}_1 < +\infty$ almost surely and when we leave $1$, either $\texttt{L}^+ \cup y^+$ occurs or $y^-$ occurs. We can now write the following decomposition of $\esp(T_1)$:
                        \[ \esp(T_1) = \proba(\texttt{L}^+_0) \espcond{T_1}{\texttt{L}^+_0} + \proba(\texttt{J}^+_0 \cup \texttt{R}^+_0) \espcond{T_1}{\texttt{J}^+_0 \cup \texttt{R}^+_0}, \]
                    and $\espcond{T_1}{\texttt{L}^+_0} = 1 + \esp(T_{1, 1})$ while
                        \begin{align*}
                            \espcond{T_1}{\texttt{J}^+_0 \cup \texttt{R}^+_0} = 1 & + \esp(T^{\pm}_1) \\ 
                            & + \probacond{L^+ \cup y^+}{\texttt{J}^+_0 \cup \texttt{R}^+_0} \esp(T_{1, 2}) \\
                            & + \probacond{y^-}{\texttt{J}^+_0 \cup \texttt{R}^+_0} \esp(T_{1, 0}),
                        \end{align*}
                    where $y^-$, $y^+$ and $\texttt{L}^+$ are the events arising at time $T^{\pm}_1$, when leaving $1$ for the first time and $T_{1,0}$, $T_{1, 1}$ and $T_{1, 2}$ are the first hitting time of $\ell + 1$ from positions $0$, $1$ and $2$, when using only \LeftBridge (see~\cref{lem-plateau} $(2)$). After plugging the different quantities using~\Cref{lem-plateau} and~\cref{fig-transition-probs} we obtain,
                        \begin{align*}
                            \esp(T_1) & = \frac{2}{3 (2n - 1)} + n(\mathcal{H}_n - \mathcal{H}_{n - \ell - 1}) \tag*{(A)}\label{(A)} \\ 
                            & = n \ln \left( \frac{1}{1 - \frac{\ell + 1}{n}} \right) + \frac{1}{2} - \frac{1}{2 \left( 1 - \frac{\ell + 1}{n}  \right)} + \o_{n \to +\infty}(1), \nonumber
                        \end{align*}
                    and applying the bounds on $\mathcal{H}_n$ from~\Cref{lem:harmonic}\footnote{Deferred in section $A$ of the appendix.} on \hyperref[(A)]{(\textsc{A})} leads to
                        \[ \esp(T_1) \leq 2(\ell + 1) \ln(2). \]
                    
                     The full proof of the theorem is provided in the appendix.
            \end{proof}
        
        \subsection{The Second Phase: Climbing the Slope} \label{par-E}

            When the second phase begins, we are in state $\ell + 1$. From here, let $T_2^1$ be the first hitting time of state $\ell + 3$ and $T_2^2$ the remaining time (before reaching position $n - \ell$ for the first time) hence $T_2 = T_2^1 + T_2^2$.
            
            Our goal here is to upper bound both $\esp(T_2^1)$ and $\esp(T_2^2)$, this is done in~\Cref{lem:phase-2-T21} and~\Cref{lemma:Learning lemma} respectively. The next lemma provides some bounds on the $Q$-table during the second phase.

            \begin{lem}[Bounds on the $Q$-Table] \label{lem:bounds}
                For any time $t \geq 0$ and state $s \in [ \ell + 1 .. n - \ell - 1 ]$, we have $Q_t[s, \texttt{J}] \geq 0$ and on states $\ell + 1 \leq s < n - \ell - 2$ (resp.  $\ell + 2 < s \leq n - \ell - 1$), the objective \RightBridge (resp. \LeftBridge) is used at most once. 
                
                Moreover, for any time $t$ during the second phase, $Q_t[0, \texttt{L}] > 0$.
            \end{lem}

            \begin{proof} (Sketch) 
                By~\Cref{lem-plateau} and~\Cref{lem:NotMuchMistakes} we have $Q_{T_1}[0, \texttt{L}] > 0$ and if $0 \leq t \leq T_1$ then $Q_t[s, \texttt{J}] = 0$ for all $s \in [ \ell + 1 .. n - \ell - 1 ]$. The result now follows by an induction on $t$, carefully considering the states $s \in \{ \ell + 2, n - \ell - 2\}$ as detailed in the appendix.
            \end{proof}


%


            \begin{lem}\label{lem:phase-2-T21}
                We have
                    \[ \esp(T_2^{1}) = \O(1). \]
            \end{lem}

            \begin{proof}
                By~\Cref{lem:bounds}, given a state $s \in \{\ell + 1, \ell + 2\}$ and $t \geq 0$ then $Q_t[s, \texttt{J}] \geq 0$. Moreover, once the objective \texttt{R} is chosen on such a state $s$ then $Q_t[s, \texttt{R}]$ becomes negative. Thus \texttt{R} is used at most once in these two states. 
                
                Now, to upper bound $\esp(T_2^{1})$, we consider the worst case in which the event $\texttt{R}^-$ occurs first (in state $\ell + 1$). On average, we stay in position $\ell$ during $\O(n / (n-\ell)) = \O(1)$ iterations (as $\ell < \frac{n}{2}$), that is, the average time until the event $\texttt{L}^+$ occurs because, by~\Cref{lem:bounds} and~\cref{lem-one-positive-per-state}, only \LeftBridge can be used in position $\ell$. Hence, after $\O(1)$ iterations on average, the state $\ell + 1$ is reached again. At this moment, only \Jump helps to leave $\ell + 1$, and as it is strictly increasing only the event $\texttt{J}^+$ allows us to escape from state $\ell + 1$. Hence, the algorithm gets stuck in $\ell + 1$ until $\texttt{J}^+$ occurs and state $\ell + 2$ is reached with an average time of $\O\left( n / (n - (\ell+1)) \right) = \O(1)$. 

                Now, define excursions which start from state $\ell + 2$ and either return to the state $\ell + 2$ without reaching $\ell + 3$ (in which case the excursion is a failure) or which succeed when state $\ell + 3$ is reached. Let $e$ (resp. $e'$) be a failing (resp. the succeeding) excursion and $\ell(e)$ (resp. $\ell(e')$) be its length, then
                \[ \esp(  T_2^{1}) = O(1) + \esp(\ell(e)) \esp(k^*) + \esp(\ell(e^\prime)), \tag*{(E)}\label{phase-2:eq1} \]
                where $k^* = \inf\{ i \geq 0 \mid e_{i+1} \textit{ is a succeeding excursion} \}$ is the number of excursions that fail. The first term in~\ref{phase-2:eq1} is the expected first hitting time of $\ell + 2$, the second one is the expected time of all failing excursions (we use \textsc{Wald}'s theorem~\cite{Wald44, DoerrK15} as these excursions are i.i.d.), and the last one is the average length of the succeeding excursion. Note that $\esp\left( \ell(e) \right) = \O(1)$ as a failing excursion is either of length $1$ (if we remain in the same state) or of length $1 + \O(1) = \O(1)$ (if we return to $\ell + 1$ at the end of the iteration). Additionally,
                \[ \esp\left( k^* \right) = \frac{1}{p_{\ell + 2 \to \ell + 3}} - 1, \]
                where $p_{\ell + 2 \to \ell + 3}$ is the transition probability from $\ell + 2$ to $\ell + 3$ and, as all objectives accept the move from $\ell + 2$ to $\ell + 3$ then, $p_{\ell + 2 \to \ell + 3}$ does not depend on the time so $p_{\ell + 2 \to \ell + 3} = \frac{n - \ell - 2}{n}$ thus $\esp \left( k^* \right) = \frac{n}{n-\ell-2} - 1 = \O(1)$, since $\ell < \frac{n}{2}$. Besides, the succeeding excursion consists in only one move, from $\ell + 2$ to $\ell + 3$, hence $\esp( \ell(e^\prime)) = 1$. Finally, combining these results leads to the bound $\esp(T_2^{1}) = \O(1)$, as desired.
            \end{proof}

        \begin{lem}[Few Visits Lemma]\label{lemma:Learning lemma}
            The algorithm visits each state $s \in [ \ell + 3 .. n - \ell - 2 ]$ at most $5$ times.
        \end{lem}
        \begin{proof}
            Recall from~\Cref{lem:bounds} that during the second phase, for any $s \in [ \ell + 1 .. n - \ell - 1 ]$ we have $Q_t[s, \texttt{J}] \geq 0$. Now, if~\cref{alg:alg1} passes from state $s \in [ \ell + 3 .. n - \ell - 1 ]$ to $s-1$ using objective $f_t$ (which is necessarily \texttt{R} or \texttt{L}) at time $t$, then the \emph{plateau penalty}
            
            is applied and 
            \begin{align*}
                Q_{t + 1}[s, f_t] &= (1 - \alpha) Q_t[s, f_t] - \alpha r + \alpha \gamma  \max_{a \in \mathscr{A}} Q_t[s - 1, a] \\
                & < \left( 1 - \alpha (1 - \gamma) - \alpha \frac{1 - \alpha + \alpha \gamma}{\alpha} \right) \frac{n - \ell - 1}{1 - \gamma} = 0,
            \end{align*}
           where the first inequality follows from~\ref{penalty-hyp}: $r > \frac{(1 - \alpha( 1 -  \gamma)) (n-\ell-1)}{\alpha (1 - \gamma)}$.
           
           Hence, $Q_{t + 1}[s, f_t] < 0 \leq Q_{t + 1}[s, \texttt{J}]$ so objective $f_t$ is never used again in $s$. This means that state $s \in [\ell + 3 .. n - \ell - 2]$ can be reached from $s+1$ only twice (using \texttt{R} or \texttt{L}) and from $s-1$ only three times (one when we first reach $s+1$ and then at most two if we eventually fall from $s$ to $s - 1$). Thus, $s$ is visited at most $5$ times as desired.
        \end{proof}

        We can now state the main result of this part.        
        \begin{thm}[Runtime of the Second Phase]\label{thm:phase2}
            We have:
                \[ \esp(T_2) \leq 5n \ln\left( \frac{n - \ell - 3}{\ell + 1} \right) + \frac{2 n}{\ell} + \O(1). \]
        \end{thm}

       \begin{proof}
           The quantity $\esp(T_2^{2})$ is the sum of the expected time spent in states $s \in  [ \ell + 3 .. n - \ell - 1 ]$ plus $2 C_1$ where $C_1$ is the upper bound we found on $\esp\left( T_2^{1} \right)$ (which accounts for the time spent in states $s < \ell + 3$ as we may fall at most twice from $\ell + 3)$, plus $1$ (the last iteration of the second phase).
           
           Now, for any state $s \in [ \ell + 3 .. n - \ell - 2 ]$, the probability to leave $p_{\text{leave}}$ to leave state $s$ is $p_{\text{leave}} \geq \frac{n-s}{n}$ so using \textsc{Wald}'s theorem, the expected time spent in state $s$ during the second phase is upper bounded by the average time we spend in $s$ before leaving it times the average number of visits to $s$ which is at most $5$ by~\Cref{lemma:Learning lemma} thus $T_s \leq \frac{5 n}{n-s}$. Besides, in state $s = n - \ell - 1$, only \LeftBridge allows to move to $s-1$ and it can be used at most once in $s$ by~\Cref{lem:bounds}. Hence from this state $s = n - \ell - 1$, as the entry $[n - \ell - 1, \texttt{J}]$ is always zero by~\Cref{lem:local-maximum}, we can fall from $s$ to $s - 1$ only once. Moreover, the probability to reach position $n - \ell$ is $p \geq \frac{\ell}{2 n}$ (consider for instance the expected time before the event $\texttt{R}^+$ occurs for the first time) so $n - \ell$ is reached in an average time less than $\frac{2 n}{\ell}$ hence
           \begin{align*}
               \esp( T_2^{2}) & \leq \sum_{i = \ell + 3}^{n - \ell - 2} \frac{5 n}{n-s} + 2 C_1  + \frac{2 n}{\ell} + \O(1)  \\
               &= 5n (\mathcal{H}_{n - \ell - 3} - \mathcal{H}_{\ell + 1}) + \frac{2 n}{\ell} + \O(1)\\ 
               & = 5n \ln\left( \frac{n - \ell - 3}{\ell + 1} \right) + \frac{2 n}{\ell} + \O(1),
           \end{align*}
           and as $T_2 = T_2^{1} + T_2^{2}$ thus $\esp\left( T_2 \right) \leq 5n \ln\left( \frac{n - \ell - 3}{\ell + 1} \right) + \frac{2 n}{\ell} + \O(1)$.
        \end{proof}
        
        \subsection{The Third Phase: Unlearn \LeftBridge} \label{par-F}
             As in the first phase, we split the third phase in two sub-phases based on the event
                \[ E_t^3 = \texttt{H}_t^n \cup \texttt{R}^+_{t, \texttt{plateau}}, \]
            i.e., \say{\textit{use {\normalfont $\texttt{R}^+$} in the plateau or hit $n$, at time $t$}}. Then, define $T_3^1$ as the first time $t \geq 0$ where $E_t^3$ occurs, if any, and $T_3^2$ the remaining time until the end of the third phase so that $T_3 = T_3^1 + T_3^2$. First, if the event $E_t^3$ never occurs then $T_3^2 = 0$, otherwise~\cref{lem-plateau} $(2)$ gives
                \[ \esp(T_3^2) \leq n (\mathcal{H}_{n - (n - \ell)} - \mathcal{H}_{n - n}) = n \mathcal{H}_{\ell} = \O(n \log(\ell)), \tag*{(B)}\label{(B)} \]
            and it remains to upper bound $\esp(T_3^1)$. To this end, the next lemma provides lower bound on both $Q_t[n - \ell - 1, \texttt{R}]$ and $Q_t[0, \texttt{R}]$ for any time $t \geq 0$. These lower bounds are useful in the study of $\esp(T_3^1)$.

            \begin{lem} \label{lem:phase-3-ineq}
                For any time $t \geq 0$, we have 
                    \[ Q_t[0,  {\normalfont \texttt{R}}] \geq - \alpha r,\ Q_t[n - \ell - 1,  {\normalfont \texttt{R}}] \geq 0, \]
                and during the third phase, from state $n - \ell - 1$ one cannot go backward.
            \end{lem}
            \begin{proof} (Sketch) Again, it is an induction on $t$, based on~\Cref{lem-plateau}, \Cref{lem:NotMuchMistakes} and~\Cref{lem:bounds}. Notably, the bound $Q_t[n - \ell - 1,  {\normalfont \texttt{R}}] \geq 0$ relies on $Q_t[0,  {\normalfont \texttt{R}}] \geq - \alpha r$ and crucially on~\ref{penalty-hyp}, especially the inequality $r < \frac{1}{\alpha \gamma}$. A detailed proof can be found in the appendix.
            \end{proof}

           According to~\Cref{lem:phase-3-ineq}, when we are in state $n - \ell - 1$, either we stay there or we move forward, right into the plateau $[n -\ell .. n - 1]$. Notably, the probability $p_{t, \mathrm{leave}}$ to leave $n - \ell - 1$ depends on the time $t$ and satisfies $p_{t, \mathrm{leave}} \geq \frac{\ell}{2 n} = \Omega( \ell / n )$ so, with an average time of $\O( n / \ell)$ we leave $n - \ell - 1$ to hit position $n - \ell$. Also, the next remark holds on the right plateau of \Jump.
            
            \begin{rem}\label{rem:phase-3}
                For any position $p \in [ n - \ell .. n - 1 ]$, almost surely either $p + 1$ is reached with an average time of $\O( n / (n - p))$ after $\texttt{R}^+$ occurred or, we leave $p$ after the event $\texttt{L}^{\pm} \cup \texttt{J}^{\pm}$ occurred in an average time of $\O(1)$.
            \end{rem}

            This remark is precious for~\cref{lem:phase-3-T31} but before expanding on it, we state the~\cref{lem:phase-3-} which gives constraints on the number of times objectives \texttt{L} and \texttt{J} can bu used in the right plateau of \Jump.
            \begin{lem} \label{lem:phase-3-}
                Consider a walk across the positions $[ n - \ell .. n - 1 ]$ of the right plateau of \Jump then, at most two transitions can be performed using objective {\normalfont \texttt{J}}, after which it cannot be used anymore in state $0$.
                
                Moreover, during the third phase, if $Q_{t_0}[0, {\normalfont \texttt{L}}] < 0$ for some $t_0 \geq 0$ then $Q_t[0, {\normalfont \texttt{L}}] < 0$ for any time $t_0 \leq t < T$.
            \end{lem}
            \begin{proof} (Sketch)
                For the first part, if one uses \texttt{J} twice in a walk $\mathcal{W}$ over the right plateau of \Jump then $Q_t[0, \texttt{J]} < -\alpha r$ and using~\Cref{lem:phase-3-ineq} we have  $-\alpha r \leq Q_t[0, \texttt{R}]$ all along the third phase thus \texttt{J} cannot be used anymore. For the second part, we use induction on $t_0 \leq t < T$. This is detailed in the appendix.
            \end{proof}

            Hence~\cref{lem:phase-3-} implies that \LeftBridge can only be chosen at most two times across the whole third phase to move from two consecutive positions of the right plateau of \Jump and after that, \texttt{L} cannot be used anymore in state $0$. Effectively once $Q_t[0, \texttt{L}] < 0$ and if \texttt{L} is selected in the plateau then $Q_{t + 1}[0, \texttt{L}] = (1 - \alpha) Q_t[0, \texttt{L}] - \alpha r + \alpha \gamma Q_t[0, \texttt{L}] < -\alpha r$ and $-\alpha r \leq Q_t[0, \texttt{R}]$ thus, \texttt{L} is never chosen again in state $0$.
            
            \begin{lem} \label{lem:phase-3-T31}
                Time $T_3^1$ satisfies
                    \[ \esp(T_3^1) = \Theta\left( \frac{n^2}{\ell^2} \right). \]
            \end{lem}
            \begin{proof} (Sketch)
                First, we upper bound the average time to go from $n - \ell - 1$ to position $n - \ell + 1 < n$ which is $\O(n^2 / \ell^2)$. Next, by considering excursions from $n - \ell + 1$ which end either when we return back to $n - \ell - 1$ (a failure) or when the event $E_t^3$ occurred (a success), we show that only a finite number of such excursions can occur and we upper bound their average length, which is a $O(n / \ell)$. Moreover, an average time of $\O(n^2 / \ell^2)$ is needed between two consecutive excursions and combining all these ingredients lead to an upper bound of $\O(n^2 / \ell^2)$ as claimed. For the lower bound, we show that there is a probability $p = \Omega(1)$ to be in $n - \ell - 1$ at time $t = T_1 + T_2 + 1$ with $Q_t[0, \texttt{R}] = -\alpha r$ and $Q_t[0, \texttt{J}] \geq 0$. Then in this setup the average time to reach position $n - \ell + 1$ is $\Omega(n^2 / \ell^2)$. A full proof is provided in the appendix.
            \end{proof}

            We are now able to state the main result of this part.
            \begin{thm}[Runtime of the Third Phase]\label{thm:phase3}
                We have:
                    \[ \esp(T_3) = \O\left( \frac{n^2}{\ell^2} + n \log(\ell) \right). \]
            \end{thm}
            \begin{proof}
                Since $T_3 = T_3^1 + T_3^2$, by \hyperref[(B)]{(\textsc{B})} and~\cref{lem:phase-3-T31} we obtain
                    \[ \esp(T_3) = \esp(T_3^1) + \esp(T_3^2) = \O\left( \frac{n^2}{\ell^2} + n \log(\ell) \right). \]
            \end{proof}
            \vspace{-0.75cm}
            
    \section{Experiments}\label{sec:exp}
        
        Besides the theoretical analysis, we also performed experiments to illustrate the efficiency of our algorithm while confirming the order of magnitude obtained for the total average runtime. We run \LRSAO on \Jump where $n$ varies from $50$ to $10000$ with a step size of $50$ and $\ell \in \{2,\ \sqrt[3]{n},\ \sqrt{n},\ \left\lfloor \frac{n - 5}{2} \right\rfloor\}$. As found during the analysis, the average runtime critically depend on the magnitude of the ratio $n / \ell$. Compared to the average runtime of~\cite{Ant1} $\O(n^2 \log(n) / \ell)$, our algorithm only needed an average runtime of $\Theta(n^2 / \ell^2 + n \log(n))$ thus, improves over previous complexity in the region $\ell = \Omega(n^\alpha)$ for any $\alpha \in [0, 1)$. Moreover, \LRSAO is \emph{optimal}\footnote{Any random local search algorithm starting from $\{0\}^n$ and using the \textsc{RandomOneBitFlip} operator needs $\Omega(n \log(n))$ calls on average to optimize \Jump.} in the region $\ell = \Theta\left( n^\alpha \right)$ for any $\alpha \in \big[ \frac{1}{2}, 1\big]$, the critical value begin $\alpha = \frac{1}{2}$ for which algorithm in~\cite{Ant1} is $\O(n^{3 / 2} \log(n))$. This justifies our choice to focus on powers of $n$ for values of $\ell$.

    
        For every pair $(n, \ell)$, we perform roughly $2000$ runs with hyper-parameters $\alpha = 0.8$, $\gamma = \frac{1}{n + 1}$ and $r = n \left( \frac{1}{\alpha (1 - \gamma)} - 1 \right)$. For small values of $\ell$, we clearly witness the average runtime being, roughly, some power $> 1$ of $n$ while for bigger values (e.g., $\ell = \Omega(\sqrt{n})$), the theoretical average runtime is $\O(n \log(n))$ and empirically, the curve grows slightly faster than $\O(n)$.
        
        Plots are in~\cref{fig:exp}, with further illustrations in the appendix.
        \begin{figure}
            \vspace*{-0.3cm}
            \hspace*{-0.5cm}
            \centering
            \begin{subfigure}{.48\linewidth}
                \centering
                \includegraphics[width=4.3cm]{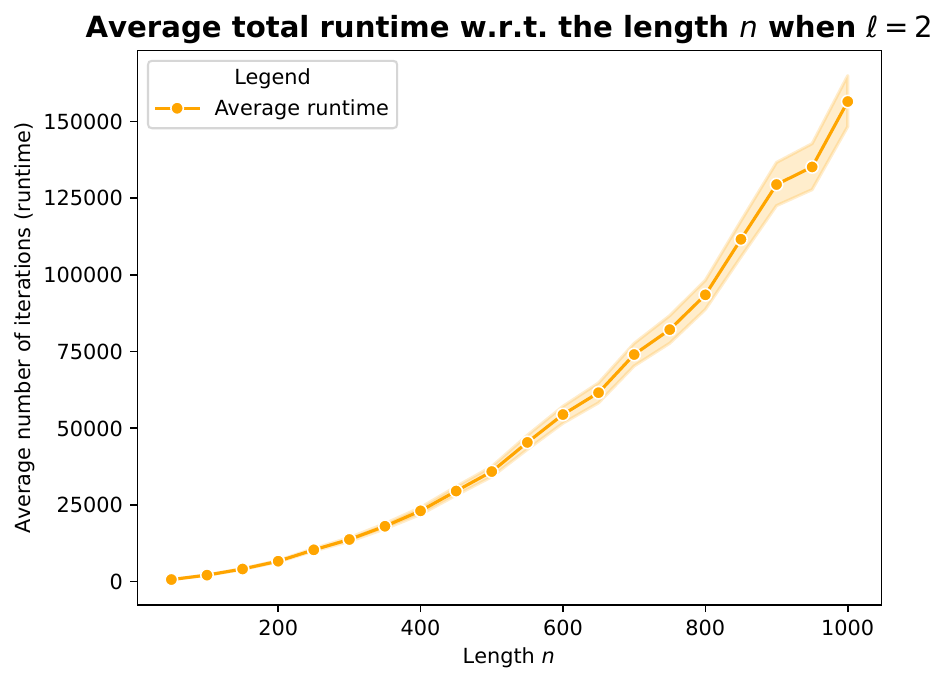}
                \caption{\footnotesize\textit{Case $\ell = 2$, $n \leq 1000$}}
            \end{subfigure}
            \hspace{0.1cm}
            \begin{subfigure}{.48\linewidth}
                \centering
                \includegraphics[width=4.3cm]{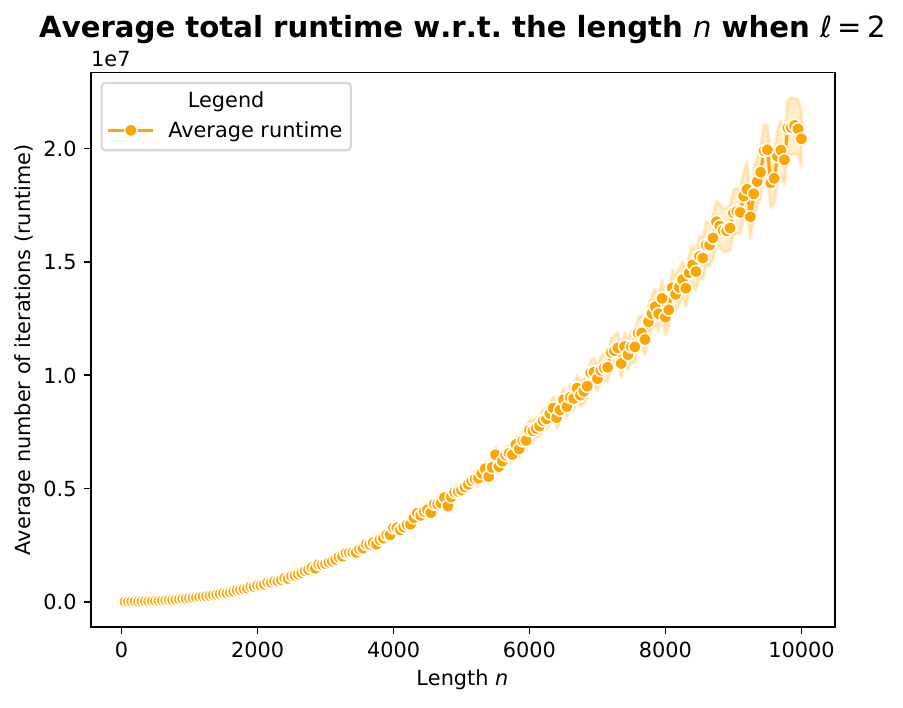}
                \caption{\footnotesize\textit{Case $\ell = 2$, $n \leq 10000$}}
            \end{subfigure}
            \\
            \hspace*{-0.5cm}
            \begin{subfigure}{.48\linewidth}
                \centering
                \includegraphics[width=4.3cm]{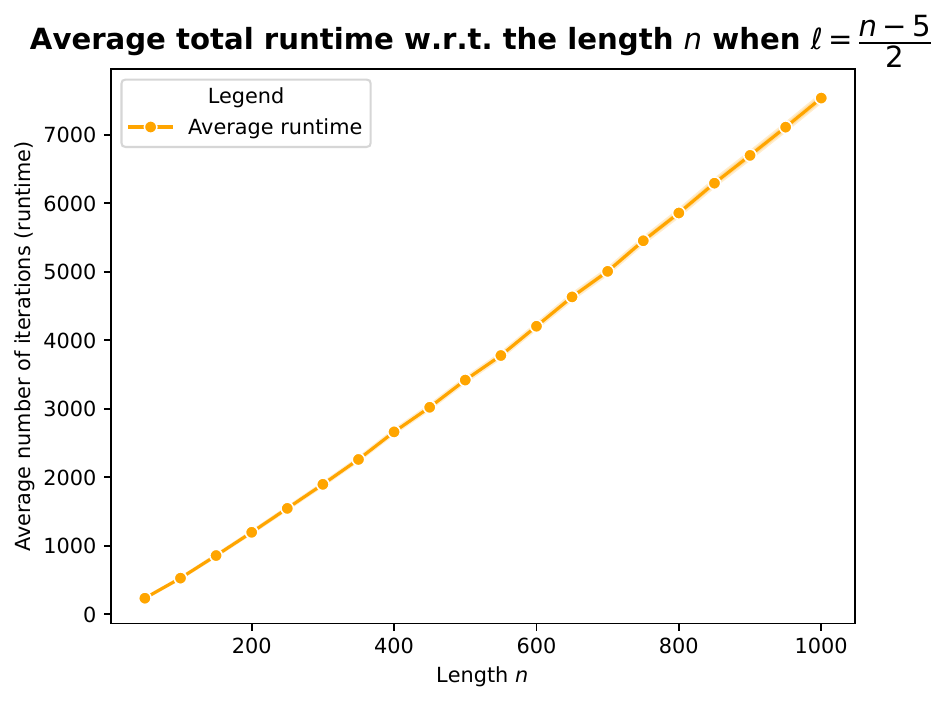}
                \caption{\footnotesize\textit{Case $\ell = \left\lfloor \frac{n - 5}{2} \right\rfloor$, $n \leq 1000$}}
            \end{subfigure}
            \hspace{0.1cm}
            \begin{subfigure}{.48\linewidth}
                \centering
                \includegraphics[width=4.3cm]{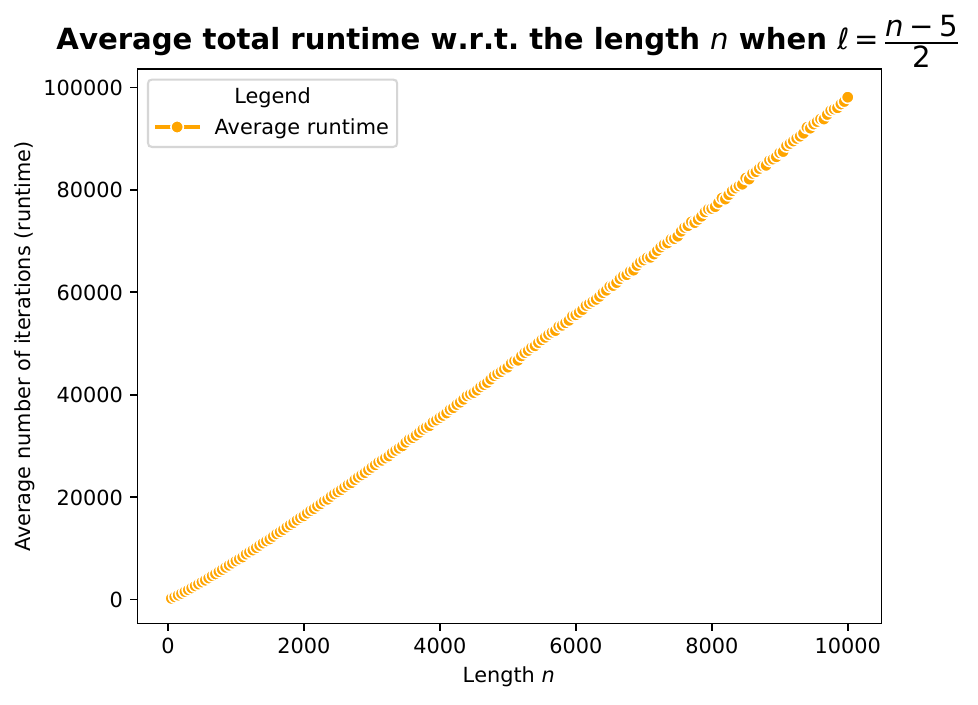}
                \caption{\footnotesize\textit{Case $\ell = \left\lfloor \frac{n - 5}{2} \right\rfloor$, $n \leq 10000$}}
            \end{subfigure}
        \caption{\small\textit{Total average runtime with $95\%$ confident intervals.}}
        \label{fig:exp}
        \vspace*{-0.8cm}
        \end{figure}

    \section{Conclusion and Future Work}
        
        By integrating an unlearning mechanism into the selection process, \LRSAO can effectively discard auxiliary objectives that are no longer beneficial in later stages of the optimization. This is achieved through a locally adaptive remuneration strategy which enables the algorithm to flexibly adjust its focus based on the changing landscape of the optimization problem. The effectiveness of \LRSAO was demonstrated on the black-box \Jump function, a difficult benchmark in evolutionary computation (\textsc{EC}). Our approach achieved a significant improvement, reducing the average runtime from $\O(n^2 \log(n) / \ell)$ attained by the \EARL~\cite{Ant1} to $\Theta(n^2 / \ell^2 + n \log(n))$. Besides this enhancement, \LRSAO does not need to be restarted. This highlights the adaptability of \LRSAO in handling non-monotonic functions. These results together with the experiments confirm the potential of \LRSAO as a promising tool for optimizing complex and dynamic problems.
    
        Future work may extend the evaluation of \LRSAO to diverse benchmarks and explore its adaptability and scalability in a broader range of optimization landscapes. Specifically, it would be interesting to study 
        benchmarks with more than three regions and hence, scenarios where either there are more than two auxiliary objectives or where the agent has to relearn old objectives (e.g., relearn \LeftBridge on a third plateau while unlearn \RightBridge). Another line of search is to explore how \LRSAO complexity is impacted when one relaxes inequalities~\ref{penalty-hyp} satisfied by penalty $r$. While we expect good performance on a larger interval, we conjecture when $r \rightarrow 0$ that \LRSAO would lose its efficiency.




\bibliographystyle{ACM-Reference-Format}
\bibliography{main, sample-base, alles_ea_master,ich_master}
\label{lastpage}

\clearpage
\appendix

\pagenumbering{Roman}

\section{Further experiments}

To compare between our algorithm and the \EARL designed in~\cite{Ant1}, we roughly perform $20000$ runs of the \EARL algorithm for $n$ from $50$ to $1000$ with a step size of $50$, as in section~\ref{sec:exp}. We use our own implementation of the \EARL as none is provided in~\cite{Ant1}. While experimenting, we found that the performances of \EARL are heavily impacted by the value of the discount factor $\gamma$ and worsen as $\gamma$ decreases. So as to guarantee a fair comparison between the two algorithms, we use the same learning rate $\alpha = 0.8$ but different discount factor: $\gamma_{\LRSAO} = \frac{1}{n + 1}$ while $\gamma_{\EARL} = 0.99$. We use the same penalty $r$ as in section~\ref{sec:exp}.

\begin{figure}
            \centering
            \begin{subfigure}{.49\linewidth}
                \centering
                \includegraphics[width=4.35cm]{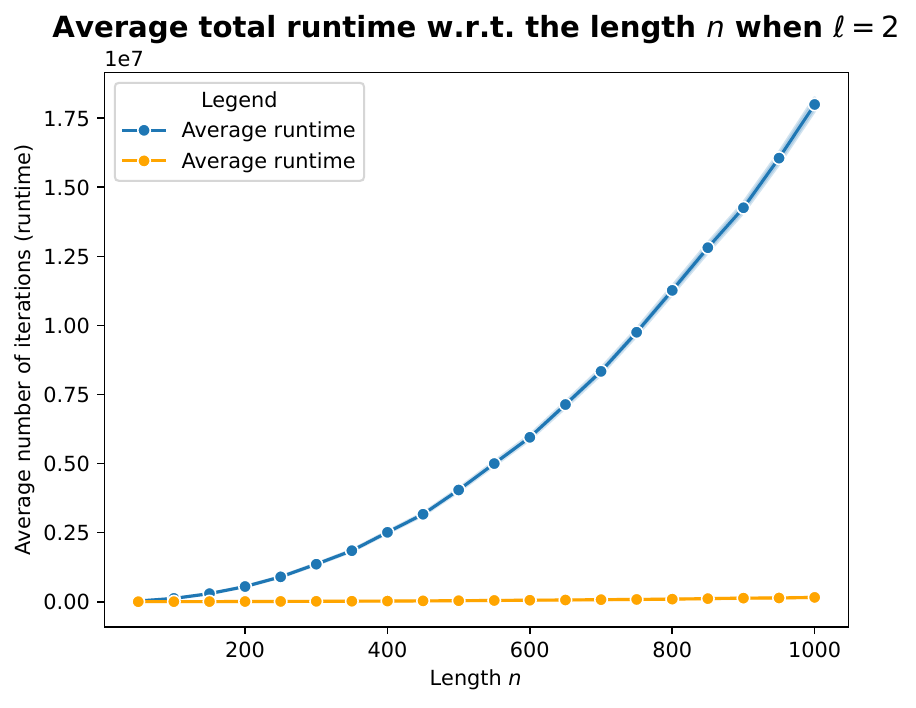}
                \caption{\footnotesize\textit{Case $\ell = 2$, $n \leq 1000$}}
            \end{subfigure}
            \hfill
            \begin{subfigure}{.49\linewidth}
                \centering
                \includegraphics[width=4.35cm]{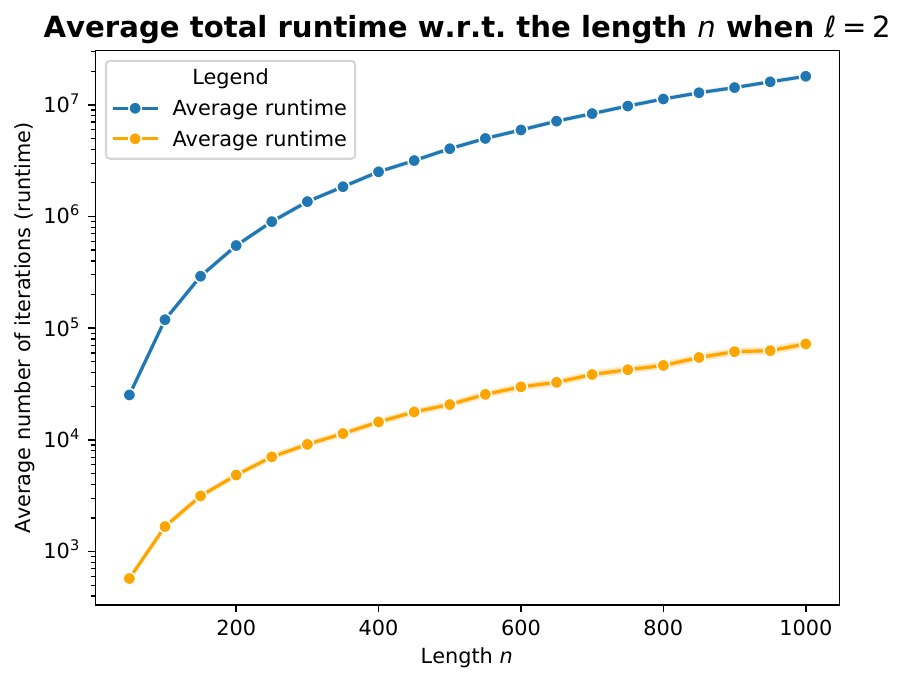}
                \caption{\footnotesize\textit{Case $\ell = 2$, $n \leq 1000$ (log scale)}}
            \end{subfigure}
            \\
            \begin{subfigure}{.49\linewidth}
                \centering
                \includegraphics[width=4.35cm]{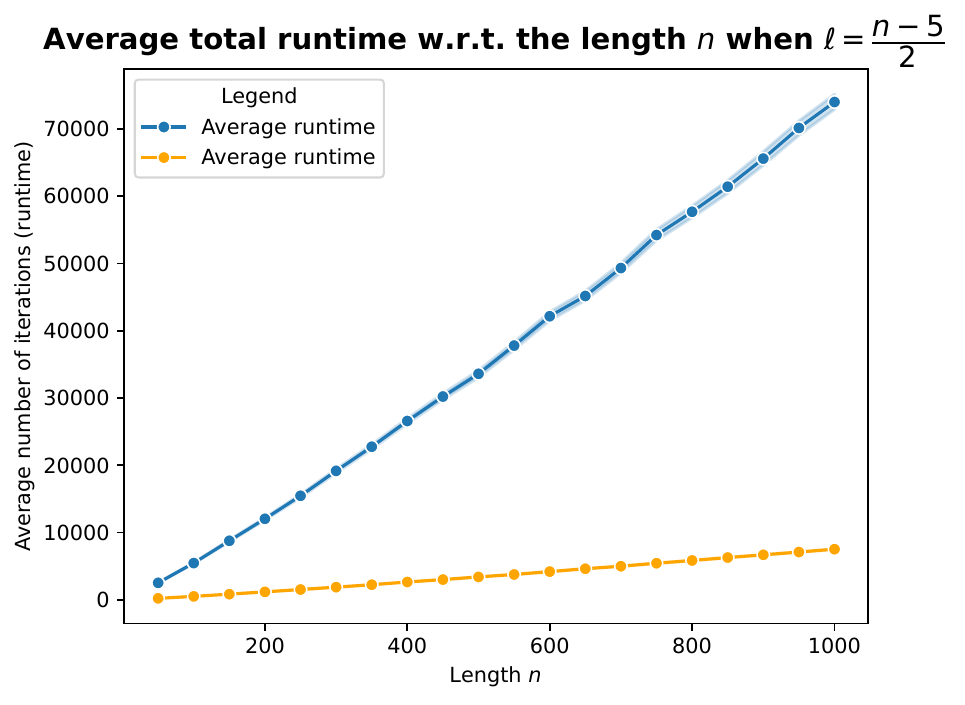}
                \caption{\footnotesize\textit{Case $\ell = \left\lfloor \frac{n - 5}{2} \right\rfloor$, $n \leq 1000$}}
            \end{subfigure}
            \hfill
            \begin{subfigure}{.49\linewidth}
                \centering
                \includegraphics[width=4.35cm]{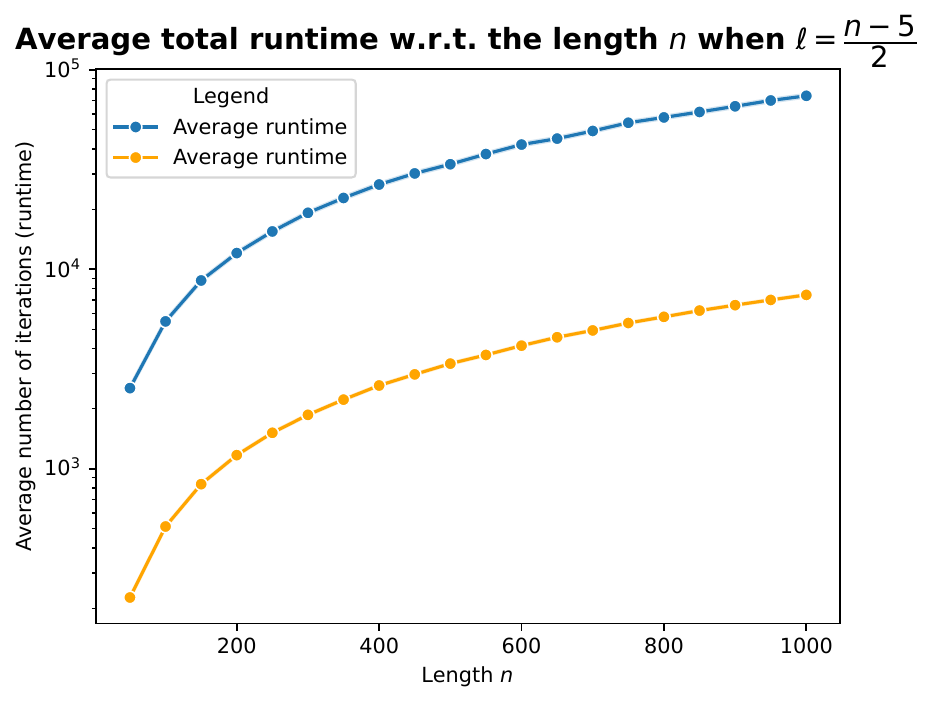}
                \caption{\footnotesize\textit{Case $\ell = \left\lfloor \frac{n - 5}{2} \right\rfloor$, $n \leq 1000$ (log scale)}}
            \end{subfigure}
        \caption{\small\textit{Total average runtime, \LRSAO (orange) vs. \EARL (blue).}}
        \label{fig:exp2}
        \end{figure}

\begin{figure}
            \centering
            \begin{subfigure}{.49\linewidth}
                \centering
                \includegraphics[width=4.35cm]{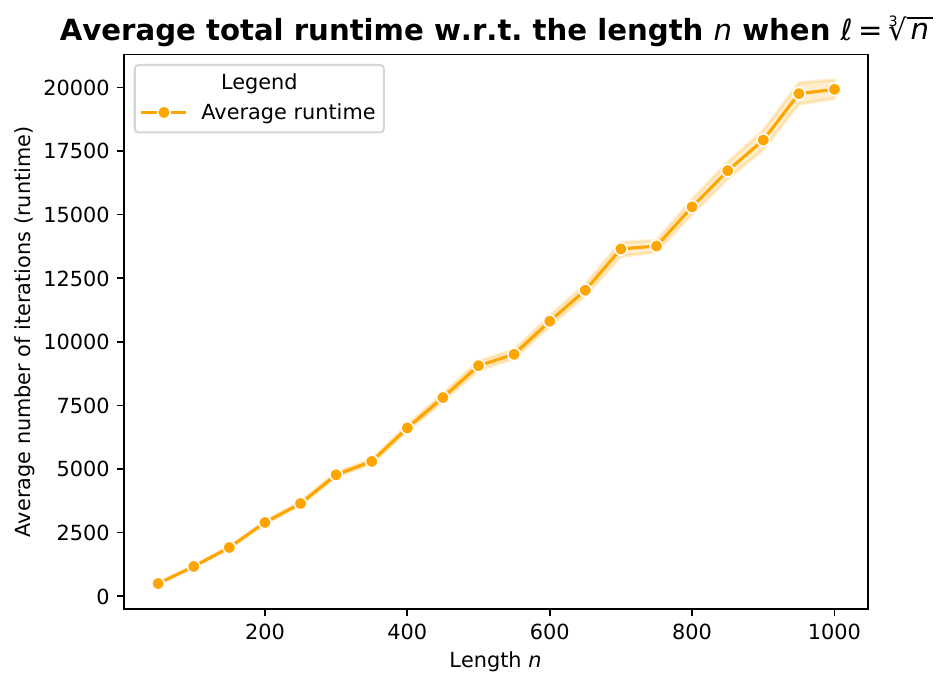}
                \caption{\footnotesize\textit{Case $\ell = \sqrt[3]{n}$, $n \leq 1000$}}
            \end{subfigure}
            \hfill
            \begin{subfigure}{.49\linewidth}
                \centering
                \includegraphics[width=4.35cm]{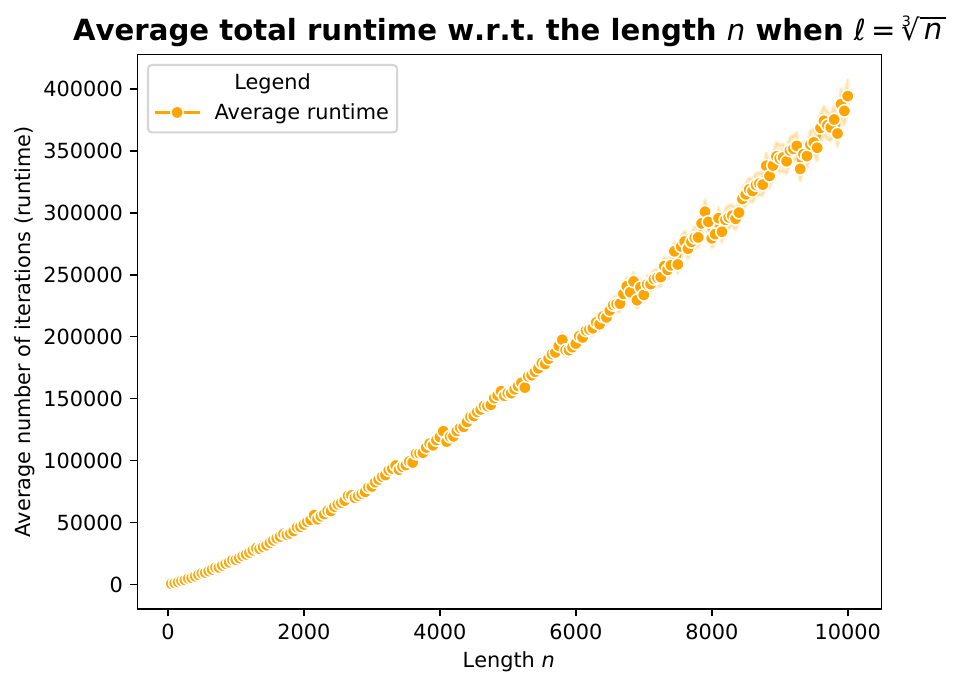}
                \caption{\footnotesize\textit{Case $\ell = \sqrt[3]{n}$, $n \leq 10000$}}
            \end{subfigure}
            \\
            \begin{subfigure}{.49\linewidth}
                \centering
                \includegraphics[width=4.35cm]{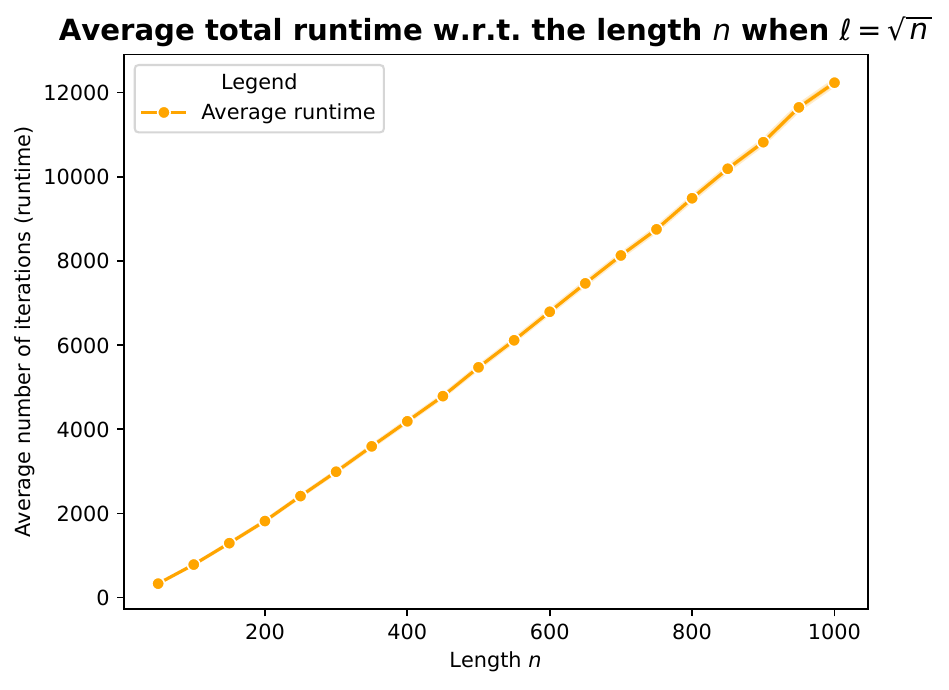}
                \caption{\footnotesize\textit{Case $\ell = \sqrt{n}$, $n \leq 1000$}}
            \end{subfigure}
            \hfill
            \begin{subfigure}{.49\linewidth}
                \centering
                \includegraphics[width=4.35cm]{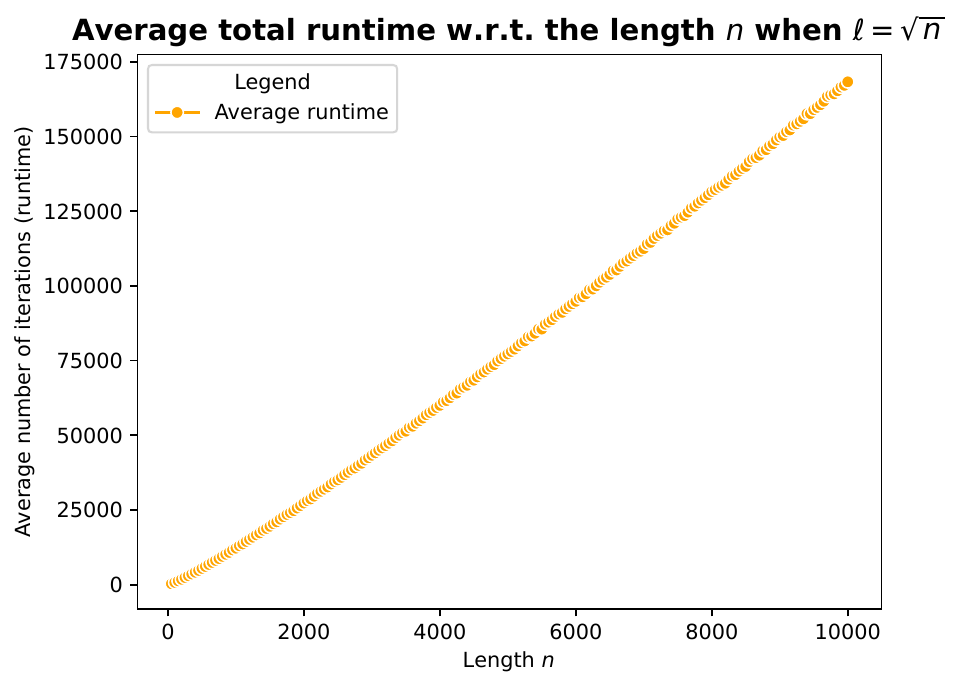}
                \caption{\footnotesize\textit{Case $\ell = \sqrt{n}$, $n \leq 10000$}}
            \end{subfigure}
        \caption{\small\textit{Total average runtime of \LRSAO for various $\ell$.}}
        \label{fig:exp3}
        \end{figure}

The plots are shown in~\cref{fig:exp2} and in~\cref{fig:exp3}.

\section{Mathematical Tools}

The following results is needed to derive good estimates on the harmonic numbers $\mathcal{H}_n = \sum_{k = 1}^n \frac{1}{k}$, $n \geq 1$. 

\begin{lem}
    The following holds for the harmonic number $\mathcal{H}_n$.\label{lem:harmonic}
    \begin{enumerate}
        \item Asymptotically, as $n \to +\infty$:
            \[ \mathcal{H}_n = \ln(n) + \gamma + \frac{1}{2 n} + \o_{n \to +\infty}\left( \frac{1}{n} \right), \]
        where $\gamma \approx 0.57721$ is the Euler–Mascheroni constant.

        \item For any positive integer $n$, we have:
            \[ \frac{1}{2 n + 1} \leq \mathcal{H}_n - \ln(n) - \gamma \leq \frac{1}{2 n}. \]
    \end{enumerate}
\end{lem}

\begin{proof}        
    For this first statement, we refer to the abundant literature where such asymptotic have been derived, e.g., \cite{olaikhan2021introduction}, \S $1.22$.

    For the second statement, given a positive integer $n$, we write $u_n = \mathcal{H}_n - \ln(n)$. By the previous statement
        \[ u_n = \mathcal{H}_n - \ln(n) \xrightarrow[n \to +\infty]{} \gamma, \]
    hence after telescoping
        \[ \sum_{k = n}^{\infty} (u_k - u_{k + 1}) = u_n - \gamma = \mathcal{H}_n - \ln(n) - \gamma. \]
    
    We now study and bound the difference $u_k - u_{k + 1}$ for a fixed integer $k \geq n$. We have
        \begin{align*}
            u_k - u_{k + 1} & = (\mathcal{H}_k - \ln(k)) - (\mathcal{H}_{k + 1} - \ln(k + 1)) \\ 
            & = \ln\left( \frac{k + 1}{k} \right) - \frac{1}{k + 1} \\ 
            & = \int_k^{k + 1} \left( \frac{1}{t} - \frac{1}{k + 1} \right) \mathrm{d} t,
        \end{align*}
    Then, to derive the upper bound, we integrate by parts as follows
        \begin{align*}
            \int_k^{k + 1} &  \left( \frac{1}{t} - \frac{1}{k + 1} \right) \mathrm{d}t \\
            & = \left[ \left( t - \frac{2 k + 1}{2} \right) \left( \frac{1}{t} - \frac{1}{k + 1} \right)  \right]_k^{k + 1} + \int_k^{k + 1} \left(t - \frac{2 k + 1}{2}  \right) \frac{\mathrm{d} t}{t^2} \\ 
            & = \left( k - \frac{2 k + 1}{2} \right) \left( \frac{1}{k} - \frac{1}{k + 1} \right) + \int_k^{k + 1} \left(t - \frac{2 k + 1}{2}  \right) \frac{\mathrm{d} t}{t^2} \\ 
            & = \frac{1}{2 k (k + 1)} - \int_k^{\frac{2 k + 1}{2}} \left(\frac{2 k + 1}{2} - t \right) \frac{\mathrm{d} t}{t^2} + \int_{\frac{2 k + 1}{2}}^{k + 1} \left(t - \frac{2 k + 1}{2} \right) \frac{\mathrm{d} t}{t^2},
        \end{align*}
    and considering the change of variable $u = 2 k + 1 - t$ in the first integral above leads to 
        \[ \int_k^{\frac{2 k + 1}{2}} \left(\frac{2 k + 1}{2} - t \right) \frac{\mathrm{d} t}{t^2} = \int_{\frac{2 k + 1}{2}}^{k + 1} \left( u - \frac{2 k + 1}{2} \right) \frac{\mathrm{d} u}{(2 k + 1 - u)^2}, \]
    hence
        \begin{align*}
            - \int_k^{\frac{2 k + 1}{2}} & \left(\frac{2 k + 1}{2} - t \right) \frac{\mathrm{d} t}{t^2} + \int_{\frac{2 k + 1}{2}}^{k + 1} \left(t - \frac{2 k + 1}{2} \right) \frac{\mathrm{d} t}{t^2} \\
            & = \int_{\frac{2 k + 1}{2}}^{k + 1} \left(t - \frac{2 k + 1}{2} \right) \left( \frac{1}{t^2} - \frac{1}{(2 k + 1 - t)^2} \right) \mathrm{d} t \\
            & \leq 0,
        \end{align*}
    since $t \geq \frac{2 k + 1}{2}$ so $t \geq 2 k + 1 - t \geq 0$ from where $\frac{1}{t^2} \leq \frac{1}{(2 k + 1 - t)^2}$. This leads to
        \[ u_k - u_{k + 1} \leq \frac{1}{2 k (k + 1)} = \frac{1}{2} \left( \frac{1}{k} - \frac{1}{k + 1} \right), \]
    and summing up these inequalities for $k \geq n$ gives the upper bound
        \[ \mathcal{H}_n - \ln(n) - \gamma \leq \frac{1}{2 n}, \]
    as desired.

    Now, for lower bound, we use a similar strategy. Let
        \[ f \colon t \mapsto \ln\left( 1 + \frac{1}{t} \right) + \ln\left( 1 + \frac{1}{t + \frac{1}{2}} \right) - \frac{1}{t + 1} - \frac{1}{t + \frac{1}{2}}, \]
    be defined on the domain $[1, +\infty)$. The function $f$ is differentiable over this domain and
        \begin{align*} 
            f'(t) & = -\frac{1}{t^2} \frac{1}{1 + \frac{1}{t}} -\frac{1}{\left( t + \frac{1}{2} \right)^2} \frac{1}{1 + \frac{1}{t + \frac{1}{2}}} + \frac{1}{(t + 1)^2} + \frac{1}{\left( t + \frac{1}{2} \right)^2} \\ 
            & = \frac{1}{(t + 1)^2} + \frac{1}{\left( t + \frac{1}{2} \right)^2} - \frac{1}{t (t + 1)} - \frac{1}{\left( t + \frac{1}{2} \right) \left( t + \frac{3}{2} \right)} \\
            & = \frac{1}{\left( t + \frac{1}{2} \right)^2 \left( t + \frac{3}{2} \right)} - \frac{1}{t (t + 1)^2} \\ 
            & = \frac{1}{t (t + 1)^2 \left( t + \frac{1}{2} \right)^2 \left( t + \frac{3}{2} \right)} \left( t (t + 1)^2 - \left( t + \frac{1}{2} \right)^2 \left( t + \frac{3}{2} \right) \right) \\ 
            & = \frac{1}{t (t + 1)^2 \left( t + \frac{1}{2} \right)^2 \left( t + \frac{3}{2} \right)} \left( t^3 + 2 t^2 + t - \left( t^3 + \frac{5}{2} t^2 + \frac{7}{4} t + \frac{3}{8} \right) \right) \\ 
            & = \frac{-1}{8 t (t + 1)^2 \left( t + \frac{1}{2} \right)^2 \left( t + \frac{3}{2} \right)} \left( 4 t^2 + 6 t + 3 \right) \\
            & < 0,
        \end{align*}
    so $f$ is decreasing (and continuous) over $[1, +\infty)$ and $f(t) \xrightarrow[t \to +\infty]{} 0$ thus, $f(t) \geq 0$ for any real number $t \geq 1$. Taking $t = k \geq 1$ gives
        \[ \int_k^{k + 1} \left( \frac{1}{t} - \frac{1}{k + 1} \right) \mathrm{d} t \geq \int_{k + \frac{1}{2}}^{k + \frac{3}{2}} \left( \frac{1}{k + \frac{1}{2}} - \frac{1}{t} \right) \mathrm{d} t, \]
    hence
        \begin{align*}
            u_k - u_{k + 1} & \geq \int_{k + \frac{1}{2}}^{k + \frac{3}{2}} \left( \frac{1}{k + \frac{1}{2}} - \frac{1}{t} \right) \mathrm{d} t \\ 
            & = \left[ (t - (k + 1)) \left( \frac{1}{k + \frac{1}{2}} - \frac{1}{t} \right) \right]_{k + \frac{1}{2}}^{k + \frac{3}{2}} + \int_{k + \frac{1}{2}}^{k + \frac{3}{2}} \frac{k + 1 - t}{t^2} \mathrm{d} t \\ 
            & = \frac{1}{2} \left( \frac{1}{k + \frac{1}{2}} - \frac{1}{k + \frac{3}{2}} \right) + \int_{k + \frac{1}{2}}^{k + 1} \frac{k + 1 - t}{t^2} \mathrm{d} t \\
            &\qquad - \int_{k + 1}^{k + \frac{3}{2}} \frac{t - (k + 1)}{t^2} \mathrm{d} t,
        \end{align*}
    and by a similar change of variable $u = 2 (k + 1) - t$ we obtain
        \[ \int_{k + \frac{1}{2}}^{k + 1} \frac{k + 1 - t}{t^2} \mathrm{d} t = \int_{k + 1}^{k + \frac{3}{2}} \frac{u - (k + 1)}{(2 (k + 1) - u)^2} \mathrm{d} t, \]
    hence
        \begin{align*}
            \int_{k + \frac{1}{2}}^{k + 1} \frac{k + 1 - t}{t^2} \mathrm{d} t & - \int_{k + 1}^{k + \frac{3}{2}} \frac{t - (k + 1)}{t^2} \mathrm{d} t \\
            & = \int_{k + 1}^{k + \frac{3}{2}} (t - (k + 1)) \left( \frac{1}{(2 (k + 1) - t)^2} - \frac{1}{t^2} \right) \mathrm{d} t \\
            & \geq 0,
        \end{align*}
    since $t \geq k + 1$ so $0 \leq 2 (k + 1) - t \leq t$ hence $\frac{1}{2 (k + 1) - t)^2} \geq \frac{1}{t^2}$. Finally,
        \[ u_k - u_{k + 1} \geq \frac{1}{2} \left( \frac{1}{k + \frac{1}{2}} - \frac{1}{k + \frac{3}{2}} \right), \]
    and summing up these inequalities yields the desired lower bound
        \[ \mathcal{H}_n - \ln(n) - \gamma \geq \frac{1}{2 n + 1}. \]

    With a closer look at the derivative of $f$, one can improve the lower bound and obtain
        \[ \frac{1}{2 n + \frac{2}{3}} \leq \mathcal{H}_n - \ln(n) - \gamma, \]
    but we do not use it in our estimations.
\end{proof}

\section{Some Properties of the $Q$-Table}

\begin{lem*}[$Q$-Table and a Local Maximum]
    For any set $\mathscr{A}$ of objectives, if state $s \in \mathcal{S}$ is a strict local maximum of an objective $a \in \mathscr{A}$ then, for any time $t \geq 0$, $Q_t[s, a] = 0$.
\end{lem*}

\begin{proof}[Proof of~\Cref{lem:local-maximum}]
  We proceed by induction on $t$. For the base case, since initially all entries of the $Q$-table are zeros at $t = 0$ then $Q_0[s, a] = 0$, as desired. Now, assume that at some time $t \geq 0$ we have $Q_t[s, a] = 0$. Then, either $s_t \neq s$ or $f_t \neq a$ so the entry $[s, a]$ is not updated during iteration $t$ thus $Q_{t + 1}[s, a] = Q_t[s, a] = 0$. Otherwise, $s_t = s$ and $f_t = a$ hence, objective $a$ is the one having (one of) the largest $Q$-value at state $s$, that is, $a \in \argmax_{a' \in \mathscr{A}} Q_t[s, a']$. In this case, as $s$ is a strict local maximum of objective $a$ and since \LRSAO always produces an offspring $x_{\texttt{new}}$ with a different position than its parent $x_t$, i.e., $\| x_{\texttt{new}} \|_1 \neq \| x_t \|_1$ thus $f_t(x_{\texttt{new}}) < f_t(x_t)$ and the move to $x_{\texttt{new}}$ is rejected so $x_{t + 1} = x_t$ and $r_{t + 1} = 0$. This leads to $s_{t + 1} = s$ and from 
  \[ Q_{t + 1}[s_t, f_t] = (1 - \alpha) Q_t[s_t, f_t] + \alpha(r_{t + 1} + \gamma \cdot \max_{a' \in \mathscr{A}} Q_t[s_{t + 1}, a']), \]
  we obtain
    \begin{align*}
       Q_{t + 1}[s, a] & = (1 - \alpha) Q_t[s, a] + \alpha(r_{t + 1} + \gamma \max_{a' \in \mathscr{A}} Q_t[s, a']) \\ 
        & = (1 - \alpha) Q_t[s, a] + \alpha \gamma Q_t[s, a] \\
        & = (1 - \alpha (1 - \gamma)) Q_t[s, a] = 0,
    \end{align*}
    since $f_t = a \in \argmax_{a' \in \mathscr{A}} Q_t[s, a']$ and $Q_t[s, a] = 0$. This achieves the proof by induction.
\end{proof}

\section{The First Phase}

\begin{lem*}[Few Mistakes Lemma]
    There exists at most one $t_{\normalfont \texttt{J}}$ and at most one $t_{\normalfont \texttt{R}}$ in $[ 0 .. T_1 - 2 ]$, such that
       \[ f_{t_{\normalfont \texttt{J}}} = {\normalfont \texttt{J}} \text{ and } f_{t_{\normalfont \texttt{R}}} = {\normalfont \texttt{R}}, \]
    and for any $0 \leq t \leq T_1$, both $Q_t[0, \texttt{J}]$ and $Q_t[0, \texttt{R}]$ lie in $\{0, -\alpha r\}$.
    
    Moreover $f_{T_1 - 1} = {\normalfont \texttt{L}}$ whenever $T_1$ is finite.
\end{lem*}
\begin{proof}[Proof of~\Cref{lem:NotMuchMistakes}]
    Using~\Cref{lem-plateau} $(1)$ and $(2)$ on the left plateau of \Jump with $t_s = 0$ then for any time $0 \leq t < T_1$, as we stay in the plateau $[0..\ell]$, we have 
    \[ Q_t[0, \texttt{L}] \geq (1 - \alpha (1 - \gamma))^{\ell(t)} Q_0[0, \texttt{L}] = 0, \]
    and if $\texttt{L}^+_{t_0, \texttt{plateau}}$ occurred at some time $0 \leq t_0 < T_1$ then we constantly choose \texttt{L} until the end of the first phase. As both \Jump and \RightBridge have a plateau in the region $[0.. \ell]$, if \texttt{J} (resp. \texttt{R}) is chosen, say for the first time, during iteration $0 \leq t < T_1 - 1$ then $Q_t[0, \texttt{J}] = 0$ since entry $[0, \texttt{J}]$ has never been updated before. Moreover, as $t + 1 < T_1$ then $\| x_{t + 1} \|_1 \in [0 .. \ell]$ so $s_t = 0 = s_{t + 1}$ and $r_{t + 1} = -r$ because $x_t$ and $x_{\texttt{new}}$ have the same fitness value. Then
        \begin{align*}
            Q_{t + 1}[0, \texttt{J}] & = (1 - \alpha) Q_t[0, \texttt{J}] - \alpha r + \alpha \gamma Q_t[0, \texttt{J}] \\ 
            & =- \alpha r < 0,
        \end{align*}
   since as $f_t = \texttt{J}$. Hence objective \texttt{J} cannot be chosen more than once and the same applies to \texttt{R}. Finally, since $\ell \geq 2$ and $x_0 = [0, \ldots, 0]$, at least $3$ steps are needed to leave the plateau $[0..\ell]$. Consequently, out of the moves from position $0$ to $1$, position $1$ to $2$ or $2$ to $3$ (which occurs almost surely), one of them must be performed using $\texttt{L}$ and so by~\cref{lem-plateau} $(2)$ \LeftBridge is used then for the rest of the walk on the plateau. In particular, the last iteration of the walk over $[0..\ell]$ must be done using \texttt{L}, that is, $f_{T_1 - 1} = \texttt{L}$. Finally, as objectives \texttt{J} and \texttt{R} cannot be used more than once then, all along the first phase, $Q_t[0, \texttt{J}]$ and $Q_t[0, \texttt{R}]$ must lie in the set $\{0, -\alpha r\}$, as desired.
\end{proof}

\begin{thm*}[Runtime of the First Phase]
    We have:
    \begin{align*}
        \esp(T_1) & = n \ln \left( \frac{1}{1 - \frac{\ell + 1}{n}} \right) + \frac{1}{2} - \frac{1}{2 \left( 1 - \frac{\ell + 1}{n}  \right)} + \o_{n \to +\infty}(1) \\ 
        & \leq 2 (\ell + 1) \ln(2).
    \end{align*}
\end{thm*}
\begin{proof}[Proof of~\Cref{thm:phase1}]
    For completeness, we recall the proof sketch of the theorem while filling in the missing details.

    The first iteration results into one of two scenarios, either the event $\texttt{L}^+_0$ occurs (and by~\cref{lem-plateau}, \texttt{L} is selected until the end of the phase) or, $\texttt{J}^+_0 \cup \texttt{R}^+_0$ occurs, say it is $x^+_0$ where $x \in \ens{\texttt{J}, \texttt{R}}$ and let $y \in \ens{\texttt{J}, \texttt{R}} \setminus \ens{x}$ be the other objective. At time $t = 1$ we are in position $1$ in both scenarios and in the second one, objective $x$ cannot be selected anymore according to~\cref{lem:NotMuchMistakes}. This leads to the transition probabilities shown in~\cref{fig-transition-probs}
        \begin{figure}
            \centering
            \resizebox{6cm}{!}{%
                \begin{tikzpicture}[->, >=stealth', auto, node distance = 4cm]
                    \node[circle, minimum size = 1.2cm, draw](0) at (0, 0) {$0$};
                    \node[circle, minimum size = 1.2cm, draw](1) at (3, 0) {$1$};
                    \node[circle, minimum size = 1.2cm, draw](2) at (6, 0) {$2$};

                    \path (1) edge [loop below] node [below] {$\frac{1}{2 n}\ (\texttt{L}^-)$} (1);
                    \path (1) edge [bend right] node [above] {$\frac{1}{2 n}\ (y^-)$} (0);
                    \path (1) edge [bend left] node [above] {$\frac{2 (n - 1)}{2n}\ (\texttt{L}^+ \cup y^+)$} (2);
                \end{tikzpicture}
            }%
            \caption{\small\textit{Transitions probabilities between $0$, $1$ and $2$ at $t = 1$.}}
            \label{fig-transition-probs-app}
            \vspace*{-0.4cm}
        \end{figure}
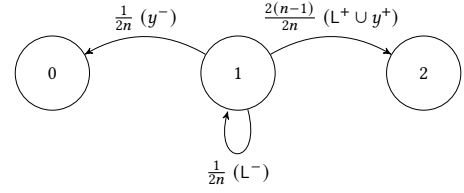
    where we remove the \textit{time index} on the events $y^\pm$ and $\texttt{L}^\pm$. Let $T^{\pm}_1 \in \N_0 \cup \ens{+\infty}$ the time taken to leave $1$, then according to~\Cref{fig-transition-probs-app}
        \[ \esp(T^{\pm}_1) = \frac{1}{1 - \frac{1}{2 n}} = \frac{2 n}{2 n - 1}, \]
    since there is a probability of $\frac{1}{2 n}$ to stay in $1$ (a failure) and $1 - \frac{1}{2 n}$ to leave (the \emph{success}). Hence, $T_1^{\pm}$ has finite expectation thus $T^{\pm}_1 < +\infty$ a.s. and when we leave position $1$, either $\texttt{L}^+ \cup y^+$ occurs or $y^-$ occurs with probability
        \[ \frac{\frac{2 (n - 1)}{2 n}}{\frac{2 (n - 1)}{2 n} + \frac{1}{2 n}} = \frac{2 (n - 1)}{2 n - 1}, \]
    for the former and
        \[  \frac{\frac{1}{2 n}}{\frac{2 (n - 1)}{2 n} + \frac{1}{2 n}} = \frac{1}{2 n - 1}, \]
    for the later. Based on these events, we can decompose $\esp(T_1)$ as
        \[ \esp(T_1) = \proba(\texttt{L}^+_0) \espcond{T_1}{\texttt{L}^+_0} + \proba(\texttt{J}^+_0 \cup \texttt{R}^+_0) \espcond{T_1}{\texttt{J}^+_0 \cup \texttt{R}^+_0}, \]
    with $\espcond{T_1}{\texttt{L}^+_0} = 1 + \esp(T_{1, 1})$ while
        \begin{align*}
            \espcond{T_1}{\texttt{J}^+_0 \cup \texttt{R}^+_0} = 1 & + \esp(T^{\pm}_1) \\ 
            & + \probacond{L^+ \cup y^+}{\texttt{J}^+_0 \cup \texttt{R}^+_0} \esp(T_{1, 2}) \\
            & + \probacond{y^-}{\texttt{J}^+_0 \cup \texttt{R}^+_0} \esp(T_{1, 0}),
        \end{align*}
    where $y^-$, $y^+$ and $\texttt{L}^+$ are the events arising at time $T^{\pm}_1$, when leaving $1$ for the first time and $T_{1,0}$, $T_{1, 1}$ and $T_{1, 2}$ are the first hitting time of $\ell + 1$ from positions $0$, $1$ and $2$, when using only \LeftBridge (see~\cref{lem-plateau} $(2)$). Now, we simply plug the value of these different quantities using~\Cref{lem-plateau} along with~\cref{fig-transition-probs-app} and some previous computations, this leads to
        \begin{align*}
            \esp(T_1) & = \frac{1}{3} \cdot \left( 1 + n (\mathcal{H}_{n - 1} - \mathcal{H}_{n - \ell - 1}) \right) \\
            &\!\  + \frac{2}{3} \left( 1 + \frac{2 n}{2 n - 1} + \frac{2 n (n - 1)}{2 n - 1} \cdot (\mathcal{H}_{n - 2} - \mathcal{H}_{n - \ell - 1})  \right. \\
            &\qquad\qquad\qquad\qquad\qquad\qquad \left. \vphantom{\frac{2 n}{2 n - 1}} + \frac{n}{2 n - 1} \cdot (\mathcal{H}_n - \mathcal{H}_{n - \ell - 1}) \right) \\ 
            & = 1 + \frac{2}{3} \cdot \left( 1 + \frac{1}{2 n - 1} \right) + n(\mathcal{H}_n - \mathcal{H}_{n - \ell - 1}) \\
            & \qquad\qquad\qquad\qquad\quad\ - \left( \frac{1}{3} + \frac{2}{3} \cdot \frac{2 n (n - 1)}{2 n - 1} \cdot \left( \frac{1}{n - 1} + \frac{1}{n} \right) \right) \\ 
            & = \frac{5}{3} + \frac{2}{3 (2n - 1)} + n(\mathcal{H}_n - \mathcal{H}_{n - \ell - 1}) - \frac{5}{3} \\ 
            & = \frac{2}{3 (2n - 1)} + n(\mathcal{H}_n - \mathcal{H}_{n - \ell - 1}),
        \end{align*}
    and the asymptotic for the harmonic numbers from~\Cref{lem:harmonic} gives
        \begin{align*}
            \esp(T_1) & = \frac{2}{3 (2n - 1)} + n(\mathcal{H}_n - \mathcal{H}_{n - \ell - 1}) \\ 
            & = \o_{n \to +\infty}(1) + n \left( \ln(n) + \gamma + \frac{1}{2 n} - \ln(n - \ell - 1) \right. \\
            &\qquad\qquad\qquad\qquad\qquad \left. \vphantom{\ln(n) + \gamma + \frac{1}{2 n} - \ln(n - \ell - 1)} -\gamma - \frac{1}{2(n - \ell - 1)} + \o_{n \to +\infty}\left( \frac{1}{n} \right) \right) \\ 
            & = n \ln\left( \frac{1}{1 - \frac{\ell + 1}{n}} \right) + \frac{1}{2} - \frac{1}{2 \left( 1 - \frac{\ell + 1}{n} \right)} + \o_{n \to +\infty}(1)
        \end{align*}
    as desired.
    
    Moreover, with the bounds on the harmonics numbers from~\Cref{lem:harmonic} and applied on $\esp(T_1) = \frac{2}{3 (2n - 1)} + n(\mathcal{H}_n - \mathcal{H}_{n - \ell - 1})$ gives
    \begin{align*}
        \esp(T_1) & = \frac{2}{3 (2n - 1)} + n(\mathcal{H}_n - \mathcal{H}_{n - \ell - 1}) \\ 
        & \leq \frac{2}{3 (2n - 1)} + n\left( \ln(n) + \gamma + \frac{1}{2 n} - \ln(n - \ell - 1) \right. \\
        &\qquad\qquad\qquad\qquad\qquad\qquad \left. \vphantom{\ln(n) + \gamma + \frac{1}{2 n} - \ln(n - \ell - 1)}- \gamma - \frac{1}{2 (n - \ell - 1) + 1} \right) \\ 
        & = n \ln\left( \frac{1}{1 - \frac{\ell + 1}{n}} \right) + \frac{1}{2} + \frac{2}{3 (2n - 1)} - \frac{n}{2 (n - \ell - 1) + 1},
    \end{align*}
    and notice that, since $3 \leq \ell + 1 \leq \left\lfloor \frac{n - 1}{2} \right\rfloor - 1 < \frac{n}{2}$ then
        \begin{align*}
            \frac{1}{2} & + \frac{2}{3 (2n - 1)} - \frac{n}{2 (n - \ell - 1) + 1} \\
            & \leq \frac{1}{2} + \frac{2}{3 (2n - 1)} - \frac{n}{2 (n - 3) + 1} \\ 
            & = \frac{1}{6 (2 n - 1) (2 n - 5)} \left( 3 (2 n - 2) (2 n - 5) + 4 (2 n - 5) - 6 n (2 n - 1) \right) \\ 
            & = \frac{1}{6 (2 n - 1) (2 n - 5)} \left( 3 (4 n^2 - 14 n + 10) + 8 n - 20 - 12 n^2 + 6 n \right) \\ 
            & = \frac{-28 n + 10}{6 (2 n - 1) (2 n - 5)} \\ 
            & < 0,
        \end{align*}
    because $n \geq 8$. Hence
    \[ \esp(T_1) \leq n \ln\left( \frac{1}{1 - \frac{\ell + 1}{n}} \right) \]
    and, considering the function $f \colon x \mapsto \ln\left( \frac{1}{1 - x} \right)$ over $\left[ 0, \frac{1}{2} \right]$, it is well-defined, and continuously differentiable, moreover
        \[ f'(x) = \frac{1}{1 - x} \text{ and } f''(x) = \frac{1}{(1 - x)^2} > 0, \]
    thus $f$ is convex on $\left[ 0, \frac{1}{2} \right]$ from where $f(x) \leq 2 x \ln(2)$ and for $x = \frac{\ell + 1}{n} \in \left[ 0, \frac{1}{2} \right]$ we obtain, finally
        \[ \esp(T_1) \leq 2 (\ell + 1) \ln(2) = 2 \ell \ln(2) + C, \]
    where $C = 2 \ln(2)$ and this concludes the proof of the theorem.
\end{proof}

\section{The Second Phase}

\begin{lem*}[Bounds on the $Q$-Table]
    For any time $t \geq 0$ and state $s \in [ \ell + 1 .. n - \ell - 1 ]$, we have $Q_t[s, \texttt{J}] \geq 0$ and on states $\ell + 1 \leq s < n - \ell - 2$ (resp.  $\ell + 2 < s \leq n - \ell - 1$), the objective \RightBridge (resp. \LeftBridge) is used at most once. 
                
    Moreover, for any time $t$ during the second phase, $Q_t[0, \texttt{L}] > 0$.
\end{lem*}

\begin{proof}[Proof of~\Cref{lem:bounds}]
    First, for the \RightBridge objective, let $t \geq 0$ and state $s \in [ \ell + 1 .. n - \ell - 3 ]$. As this objective has a plateau over $[0, n - \ell - 2]$, if the event $\texttt{R}^{\pm}_t$ occurs at some time $t \geq 0$ such that $s_t = s$ then $r_{t + 1} = -r$ and 
        \begin{align*}
            Q_{t + 1}[s, \texttt{R}] & = (1 - \alpha) Q_t[s, \texttt{R}] -\alpha r + \alpha \gamma \max_{a \in \mathscr{A}} Q_t[s_{t + 1}, a] \\ 
            & < (1 - \alpha) \frac{n - \ell - 1}{1 - \gamma} -\alpha r + \alpha \gamma \frac{n - \ell - 1}{1 - \gamma} \\ 
            & = (1 - \alpha(1 - \gamma)) \frac{n - \ell - 1}{1 - \gamma} - \alpha r \\ 
            & \leq 0,
        \end{align*}
    where we use~\Cref{lemma:inequalities} to upper bound the entries of the $Q$-table and inequalities~\ref{penalty-hyp} to deduce the last line. Similarly for \LeftBridge, if we consider some state $s \in [ \ell + 3 .. n - \ell - 1 ]$, as this objective has a plateau over $[\ell + 2, n]$ then, if the event $\texttt{L}^{\pm}_t$ occurs at time $t \geq 0$ such that $s_t = s$ then again $r_{t + 1} = -r$ and the same computation as before leads to $Q_{t + 1}[s, \texttt{L}] < 0$. This proves that objectives \texttt{L} and \texttt{R} are used at most once, as desired.

    Now, it is enough to ensure that for any time $t \geq 0$ and any state $s \in [ \ell + 1 .. n - \ell - 1 ]$ we have $Q_t[s, \texttt{J}] \geq 0$ and this will prove the first statement of the lemma. We use induction on the time $t \geq 0$ to prove that $Q_t[s, \texttt{J}] \geq 0$ for any state $s \in [ \ell + 1 .. n - \ell - 1 ]$. For the base case, since initially at time $t = 0$ all the entries of the $Q$-table are set to zero and because during all the first phase, we stay in the left plateau of \Jump, then none of the entries $[s, \texttt{J}]$ have been updated for any $s \in [\ell + 1 .. n - \ell - 1]$ hence, for any $0 \leq t < T_1$, we have $Q_t[s, \texttt{J}] = 0$. Moreover, as in time $T_1$ we are precisely in state $\ell + 1$ for the first time, we still have $Q_{T_1}[s, \texttt{J}] = 0$ as none of the entries $[\ell + 1, \cdot]$ have been updated at time $T_1$, when we first reach state $\ell + 1$. Now, assume at some time $t \geq T_1$ that $Q_t[s, \texttt{J}] \geq 0$ for all states $s \in [ \ell + 1 .. n - \ell - 1 ]$, then during iteration $t$, either $f_t \neq \texttt{J}$ in which case $Q_{t + 1}[\cdot, \texttt{J}] = Q_t[\cdot, \texttt{J}]$, i.e., entries of the $Q$-table for \texttt{J} are unchanged and the inequalities on all states $s \in [\ell + 1 .. n - \ell - 1]$ still hold. Otherwise, if $f_t = \texttt{J}$, we then distinguish between the events $\texttt{J}^+_t$ and $\texttt{J}^-_t$. In the case  $\texttt{J}^-_t$ occurs, since \Jump is \emph{strictly increasing} in the region $[\ell + 1 .. n - \ell - 1]$ then the move to $x_{\texttt{new}}$ is rejected and we stay in the same position, that is, $s_{t + 1} = s$ and $r_{t + 1} = 0$ hence
        \begin{align*}
            Q_{t + 1}[s, \texttt{J}] & = (1 - \alpha) Q_t[s, \texttt{J}] + \alpha \gamma Q_t[s, \texttt{J}] \\ 
            & = (1 - \alpha (1 - \gamma)) Q_t[s, \texttt{J}] \\ 
            & \geq 0,
        \end{align*}
    where we used the induction hypothesis. Hence, $Q_{t + 1}[s, \texttt{J}] \geq 0$. Now, if the event $\texttt{J}^+_t$ occurs instead then $s_{t + 1} = s + 1$ and $r_{t + 1} = 1$ except in the case where $s = n - \ell - 1$ where $s_{t + 1} = s$ and $r_{t + 1} = 0$ hence, taking care of that, we obtain
        \begin{align*}
            Q_{t + 1}[s, \texttt{J}] & = (1 - \alpha) Q_t[s, \texttt{J}] + \alpha r_{t + 1} + \alpha \gamma \max_{a \in \mathscr{A}} Q_t[s_{t + 1}, a] \\ 
            & \geq \alpha \gamma \max_{a \in \mathscr{A}} Q_t[s_{t + 1}, a] \\ 
            & \geq \alpha \gamma Q_t[s_{t + 1}, \texttt{J}] \\
            & \geq 0,
        \end{align*}
    where we used the induction hypothesis as $\ell + 1 \leq s_{t + 1} \leq n - \ell - 1$. Hence, inequalities $Q_{t + 1}[s, \texttt{J}]$ hold for any state $s \in [ \ell + 1 .. n - \ell - 1 ]$ and this achieves the induction and the first part of the lemma.
    
    For the last part of the statement, by~\Cref{lem-plateau} $(1)$, for any time $0 \leq t < T_1$ we have $Q_t[0, \texttt{L}] \geq 0$ and using~\Cref{lem:NotMuchMistakes}, as $f_{T_1 - 1} = \texttt{L}$, it is \LeftBridge which is used at the end of the first phase and as it is strictly increasing from position $\ell$ to $\ell + 1$, we obtain $r_{t + 1} = 1$ hence
        \begin{align*}
            Q_{T_1}[0, \texttt{L}] & = (1 - \alpha) Q_{T_1 - 1}[0, \texttt{L}] + \alpha + \alpha \gamma \max_{a \in \mathscr{A}} Q_{T_1 - 1}[s_{T_1}, a] \\ 
            & = (1 - \alpha) Q_{T_1 - 1}[0, \texttt{L}] + \alpha \\ 
            & > 0,
        \end{align*}
    because $s_{T_1} = \ell + 1$ and $Q_{T_1 - 1}[\ell + 1, \cdot] = 0$ as this state has not been visited before. This proves that $Q_{T_1}[0, \texttt{L}] > 0$. Assume now we have $Q_t[0, \texttt{L}] > 0$ for some time $T_1 \leq t < T_1 + T_2$, that is, during the second phase. Then, since the second phase precisely ends when position $n - \ell$ is reached for the first time, the only way to update the entry $[0, \cdot]$ in the $Q$-table during the second phase is to hit the left plateau of \Jump, and more precisely, to hit position $\ell$ at least. Using~\Cref{lem-one-positive-per-state} and since $Q_t[0, \texttt{L}] > 0$ by the induction hypothesis, the first time $t_0$ (if any) when we reached $\ell$ during the second phase, we necessarily selects \LeftBridge. That being said, at time $t$ either $s_t \neq 0$ in which case, whatever $f_t \neq \texttt{L}$ or $f_t = \texttt{L}$, the entry $[0, \texttt{L}]$ is not updated so it stays positive. Otherwise, if $f_t = \texttt{L}$ and $s_t = 0$ then, as \LeftBridge is strictly increasing in the region $[0.. \ell + 1]$ then $r_{t + 1} \in \{0, 1\}$ (depending on whenever the event $\texttt{L}_t^+$ or $\texttt{L}_t^-$ occurs) and we have
        \begin{align*}
            Q_{t + 1}[0, \texttt{L}] & = (1 - \alpha) Q_t[0, \texttt{L}] + \alpha r_{t + 1} + \alpha \gamma \max_{a \in \mathscr{A}} Q_t[s_{t + 1}, a] \\ 
            & \geq  (1 - \alpha) Q_t[0, \texttt{L}]  \\ 
            & > 0,
        \end{align*}
    because $s_{t + 1} \in \{0, \ell + 1\}$ and $Q_t[\ell + 1, \texttt{J}] \geq 0$ as we proved earlier. So $\max_{a \in \mathscr{A}} Q_t[s_{t + 1}, a] \geq 0$ which achieves the proof of the lemma.
\end{proof}

\section{The Third Phase}
\begin{lem*}
    For any time $t \geq 0$, we have 
        \[ Q_t[0,  {\normalfont \texttt{R}}] \geq - \alpha r,\ Q_t[n - \ell - 1,  {\normalfont \texttt{R}}] \geq 0, \]
    and during the third phase, from state $n - \ell - 1$ one cannot go backward.
\end{lem*}

\begin{proof}[Proof of~\Cref{lem:phase-3-ineq}]
    First, lets us show the two inequalities hold during the first and second phase. Then, we use induction to prove that these three properties still hold during the third phase. By~\Cref{lem:NotMuchMistakes}, we know that, during the first phase, $Q_t[0, \texttt{R}] \in \{-\alpha r, 0\}$ and $Q_t[n - \ell - 1, \texttt{R}] = 0$. Moreover, by~\Cref{lem:bounds}, since $Q_t[0, \texttt{L}] > 0$ during all the second phase, we conclude that \RightBridge is never selected when we come back to position $\ell$ (so in the left plateau of \Jump) thus $Q_t[0, \texttt{R}]$ is unchanged during the second phase. For the entry $[n - \ell - 1, \texttt{R}]$, we proceed by induction on $t$. We have already shown that it is non-negative at the beginning of the second phase. Now still during the second phase, assume $Q_t[n - \ell - 1, \texttt{R}] \geq 0$ then, either $f_t \neq \texttt{R}$ or $s_t \neq n - \ell - 1$ in which case the entry $[n - \ell - 1, \texttt{R}]$ is unchanged, i.e., still zero. Otherwise, if $f_t = \texttt{R}$ and $s_t = n - \ell - 1$ then, as \RightBridge is increasing over $[n - \ell - 1, n]$, we have $s_{t + 1} \in \{n - \ell - 1, 0\}$ and $r_{t + 1} \in \{0, 1\}$ hence
        \begin{align*}
            Q_{t + 1}[n - \ell - 1, \texttt{R}] & = (1 - \alpha) Q_t[n - \ell - 1, \texttt{R}] + \alpha r_{t + 1} \\
            &\qquad\qquad+ \alpha \gamma \max_{a \in \mathscr{A}} Q_t[s_{t + 1}, a] \\ 
            & \geq (1 - \alpha) Q_t[n - \ell - 1, \texttt{R}] \\
            & \geq 0,
        \end{align*}
    because either $s_{t + 1} = 0$ hence $\max_{a \in \mathscr{A}} Q_t[s_{t + 1}, a] = Q_t[0, \texttt{L}] > 0$ by~\Cref{lem:bounds} or, $s_{t + 1} = n - \ell - 1$ from where $\max_{a \in \mathscr{A}} Q_t[s_{t + 1}, a] = Q_t[n - \ell - 1, \texttt{R}] \geq 0$. Thus, the quantity $Q_{t + 1}[n - \ell - 1, \texttt{R}]$ stays non-negative during all the second phase.

    We are now at the beginning of the third phase, we will show for any time $t \geq 0$ in this phase that $Q_t[0, \texttt{R}] \geq -\alpha r$, $Q_t[n - \ell - 1, \texttt{R}] \geq 0$, $Q_t[n - \ell - 1, \texttt{R}] > Q_t[n - \ell - 1, \texttt{L}]$ and $\| x_t \|_1 \geq n - \ell - 1$ (hence, we cannot go beyond state $n - \ell - 1$ anymore). Since at the end of the second phase we have $\| x_t \|_1 = n - \ell$ then, either \RightBridge or \LeftBridge was used during the second phase to move from $n - \ell - 1$ to $n - \ell$. If \LeftBridge was used, say at time $t$, then we would have $s_{t + 1} = 0$, $r_{t + 1} = -r$ and
        \begin{align*}
            Q_{t + 1}[n - \ell - 1, \texttt{L}] & = (1 - \alpha) Q_t[n - \ell - 1, \texttt{L}] -\alpha r + \alpha \gamma \max_{a \in \mathscr{A}} Q_t[0, a] \\ 
            & < (1 - \alpha) \frac{n - \ell - 1}{1 - \gamma} -\alpha r + \alpha \gamma \frac{n - \ell - 1}{1 - \gamma} \\ 
            & = (1 - \alpha(1 - \gamma)) \frac{n - \ell - 1}{1 - \gamma} - \alpha r \\ 
            & \leq 0,
        \end{align*}
    where we use~\Cref{lemma:inequalities} to upper bound the entries of the $Q$-table along with inequalities~\ref{penalty-hyp}. Hence, $Q_t[n - \ell - 1, \texttt{L}] < 0 \leq Q_t[n - \ell - 1, \texttt{R}]$, as desired (recall that we have shown before the inequality $0 \leq Q_t[n - \ell - 1, \texttt{R}]$ during the second phase). However now, if \RightBridge was used instead then $s_{t + 1} = 0$, $r_{t + 1} = 1$ and
        \begin{align*}
            Q_{t + 1}[n - \ell - 1, \texttt{R}] & = (1 - \alpha) Q_t[n - \ell - 1, \texttt{R}] + \alpha + \alpha \gamma \max_{a \in \mathscr{A}} Q_t[0, a] \\ 
            & \geq \alpha + Q_t[0, \texttt{L}] \\
            & > 0,
        \end{align*}
    as inequalities $0 \leq Q_t[n - \ell - 1, \texttt{R}]$ and $Q_t[0, \texttt{L}]$ hold during the second phase as shown previously and in~\Cref{lem:bounds}. Moreover, as proved in the previous paragraph, $Q_t[0, \texttt{R}] \geq -\alpha r$ holds initially at the beginning of the third phase.
    
    Now, assume these four properties hold at some time $t < T$ during the third phase. We distinguish two cases, either $\| x_t\|_1 \in [ n - \ell .. n - 1 ]$, i.e., we are in the right plateau of \Jump so we still have $\| x_{t + 1} \|_1 \in [ n - \ell - 1 .. n ]$. Moreover, in that case, the entry $[n - \ell - 1, \cdot]$ is not updated so we still have $Q_{t + 1}[n - \ell - 1, \texttt{R}] \geq 0$ and $Q_{t + 1}[n - \ell - 1, \texttt{R}] > Q_{t + 1}[n - \ell - 1, \texttt{L}]$ and, by~\Cref{lem-plateau}, since we are in a plateau of \RightBridge, if $\| x_{t + 1} \|_1 \neq n$ and this objective is selected then 
        \[ Q_{t + 1}[0, \texttt{R}] \geq (1 - \alpha (1 - \gamma)) Q_t[0, \texttt{R}] \geq \min\{ 0, Q_t[0, \texttt{R}] \} \geq -\alpha r, \]
    or, if $x_{t + 1} = [1, \ldots, 1]$ then $r_{t + 1} = 1$ hence
        \[ Q_{t + 1}[0, \texttt{R}] \geq \alpha + (1 - \alpha) Q_t[0, \texttt{R}] > -\alpha r, \]
    since all entries $[n, \cdot]$ are zero when we first reach state $n$. Thus all the four properties hold. Next, consider the other case when $\| x_t \|_1 = n - \ell - 1$. There, the entries $[0, \cdot]$ are not updated and, since $Q_t[n - \ell - 1, \texttt{R}] > Q_t[n - \ell - 1, \texttt{L}]$ and $Q_t[n - \ell - 1, \texttt{R}] \geq 0 = Q_t[n - \ell - 1, \texttt{J}]$ (see~\Cref{lem:local-maximum}) we deduce that \texttt{L} cannot be used anymore hence $\| x_{t + 1} \|_1 \in \{n - \ell - 1, n - \ell\}$ as \Jump and \RightBridge reject moves away of the global maximum $x^*$. Thus, it only remains to show that we still have both $Q_{t + 1}[n - \ell - 1, \texttt{R}] > Q_{t + 1}[n - \ell - 1, \texttt{L}]$ and $Q_{t + 1}[n - \ell - 1, \texttt{R}] \geq 0$. Of course, we can only have $f_t \in \{\texttt{J}, \texttt{R}\}$ and if $f_t \neq \texttt{R}$ then the entries $[n - \ell - 1, \texttt{R}]$ and $[n - \ell - 1, \texttt{L}]$ are unchanged so the inequalities are still fulfilled. Otherwise, if $f_t = \texttt{R}$ then either $s_{t + 1} = n - \ell - 1$ so $r_{t + 1} = 0$ hence
        \[ Q_{t + 1}[n - \ell - 1, \texttt{R}] = (1 - \alpha(1 - \gamma)) Q_t[n - \ell - 1, \texttt{R}], \]
    and, either $Q_t[n - \ell - 1, \texttt{R}] = 0$ so $Q_{t + 1}[n - \ell - 1, \texttt{R}] = 0 > Q_{t + 1}[n - \ell - 1, \texttt{L}]$ or, $Q_t[n - \ell - 1, \texttt{R}] > 0$ in which case $Q_{t + 1}[n - \ell - 1, \texttt{R}] > 0$ and by~\Cref{lem-one-positive-per-state}, only one objective can have a positive entry in state $n - \ell - 1$ at time $t + 1$ thus 
        \[ Q_{t + 1}[n - \ell - 1, \texttt{R}] > 0 \geq Q_{t + 1}[n - \ell - 1, \texttt{L}], \]
    which gives the desired inequalities. On the other hand, if $s_{t + 1} = 0$ then $r_{t + 1} = 1$ and 
        \begin{align*}
            Q_{t + 1}[n - \ell - 1, \texttt{R}] & = (1 - \alpha) Q_t[n - \ell - 1, \texttt{R}] + \alpha + \alpha \gamma \max_{a \in \mathscr{A}} Q_t[0, a] \\ 
            & \geq (1 - \alpha) Q_t[n - \ell - 1, \texttt{R}] + \alpha - \alpha^2 \gamma r \\ 
            & = (1 - \alpha) Q_t[n - \ell - 1, \texttt{R}] + \alpha (1  - \alpha \gamma r) \\ 
            & > (1 - \alpha) Q_t[n - \ell - 1, \texttt{R}],
        \end{align*}
    since, $r < \frac{1}{\alpha \gamma}$ by~\ref{penalty-hyp} and, using $Q_t[0, \texttt{R}] \geq -\alpha r$ from the induction hypothesis, we obtain $\max_{a \in \mathscr{A}} Q_t[0, a] \geq -\alpha r$. By distinguishing the cases $Q_t[n - \ell - 1, \texttt{R}] = 0$ and $Q_t[n - \ell - 1, \texttt{R}] > 0$ (and using~\Cref{lem:local-maximum}) we also obtain the two desired inequalities $Q_{t + 1}[n - \ell - 1, \texttt{R}] > Q_{t + 1}[n - \ell - 1, \texttt{L}]$ and $Q_{t + 1}[n - \ell - 1, \texttt{R}] \geq 0$. This concludes the proof of the lemma.
\end{proof}

\begin{lem*}
    Consider a walk across the positions $[ n - \ell .. n - 1 ]$ of the right plateau of \Jump then, at most two transitions can be performed using objective {\normalfont \texttt{J}}, after which it cannot be used anymore in state $0$.
                
    Moreover, during the third phase, if $Q_{t_0}[0, {\normalfont \texttt{L}}] < 0$ for some $t_0 \geq 0$ then $Q_t[0, {\normalfont \texttt{L}}] < 0$ for any time $t_0 \leq t < T$.
\end{lem*}

\begin{proof}[Proof of~\Cref{lem:phase-3-}]
    For the first part of the statement, consider a walk $\mathcal{W}$ on the right plateau of \Jump, i.e., over positions $[n - \ell.. n- 1]$. For the sake of contradiction, assume \texttt{J} has been selected three times or more during the walk $\mathcal{W}$. As we do not leave the plateau, all transitions made are between positions of this plateau thus, every time \Jump is used, a penalty of $-r$ is given to the entry $[0, \texttt{J}]$. Recall that, by~\Cref{lem:phase-3-ineq}, for any time $t \geq 0$, we have $Q_t[0, \texttt{R}] \geq -\alpha r$. Now, consider the first three times $t_1 < t_2 < t_3 < T$ of the walk $\mathcal{W}$ where \texttt{J} was used then, 
        \begin{align*}
            Q_{t_1 + 1}[0, \texttt{J}] & = (1 - \alpha) Q_{t_1}[0, \texttt{J}] - \alpha r + \alpha \gamma Q_{t_1}[0, \texttt{J}] \\
            & = (1 - \alpha (1 - \gamma)) Q_{t_1}[0, \texttt{J}] - \alpha r \\ 
            & < (1 - \alpha (1 - \gamma)) \frac{n - \ell - 1}{1 - \gamma} - \alpha r \\ 
            & \leq 0,
        \end{align*}
    where we use~\Cref{lemma:inequalities} and inequalities~\ref{penalty-hyp}. Hence, after using \Jump for the first time $Q_{t_1 + 1}[0, \texttt{J}] < 0$. Now, after using \Jump for the second time, we have $Q_{t_2}[0, \texttt{J}] = Q_{t_1 + 1}[0, \texttt{J}] < 0$ as this entry has not been updated so far, and
        \begin{align*}
            Q_{t_2 + 1}[0, \texttt{J}] & = (1 - \alpha) Q_{t_2}[0, \texttt{J}] - \alpha r + \alpha \gamma Q_{t_2}[0, \texttt{J}] \\
            & = (1 - \alpha (1 - \gamma)) Q_{t_1 + 1}[0, \texttt{J}] - \alpha r \\ 
            & < - \alpha r,
        \end{align*}
    but now, $Q_{t_2 + 1}[0, \texttt{J}] = Q_{t_3}[0, \texttt{J}] < -\alpha r \leq Q_t[0, \texttt{R}]$ which is absurd: objective \texttt{R} should have been preferred over \texttt{J} at time $t_3$. This is incompatible with the behavior of~\Cref{alg:alg1} thus, \Jump is selected at most twice during such a walk. Notably, if \Jump is effectively selected two times during a walk $\mathcal{W}$ in the plateau then it cannot be selected anymore in any position $[n - \ell.. n - 1]$ since, by~\Cref{lem:phase-3-ineq}, we always have $Q_t[0, \texttt{R}] \geq -\alpha r$.

    On the other hand, for the second part of the lemma, we prove it by induction on $t$. Assume $Q_{t_0}[0, \texttt{L}] < 0$ for some time $t_0 \geq 0$ during the third phase then, $Q_t[0, \texttt{L}] < 0$ holds at time $t = t_0$. Now, assume $Q_t[0, \texttt{L}] < 0$ holds for some time $t_0 \leq t < T - 1$ then, the only way entry $[0, \texttt{L}]$ is updated during iteration $t$ is to select \LeftBridge while being in the right plateau of \Jump hence, assume $s_t = 0$ and $f_t = \texttt{L}$ thus, as $t < T - 1$ then $S_{t + 1} \neq n$ and since we are in a plateau of \LeftBridge, the reward is $r_{t + 1} = -r$ thus
        \begin{align*}
            Q_{t + 1}[0, \texttt{L}] & = (1 - \alpha) Q_t[0, \texttt{L}] - \alpha r + \alpha \gamma \max_{a \in \mathscr{A}} Q_t[s_{t + 1}, a] \\ 
            & < (1 - \alpha(1 - \gamma)) \frac{n - \ell - 1}{1 - \gamma} - \alpha r \\
            & \leq 0,
        \end{align*}
    where we use, again, \Cref{lemma:inequalities} and the assumptions~\ref{penalty-hyp} to upper bound both $\max_{a \in \mathscr{A}} Q_t[s_{t + 1}, a]$ and $Q_t[0, \texttt{L}]$. This shows that $Q_{t + 1}[0, \texttt{L}] < 0$ still holds, as desired, which achieves the proof.
\end{proof}

\begin{lem*}
    Time $T_3^1$ satisfies
        \[ \esp(T_3^1) = \Theta\left( \frac{n^2}{\ell^2} \right). \]
\end{lem*}
\begin{proof}[Proof of~\cref{lem:phase-3-T31}]
    We start to derive the upper bound. First, we upper bound the average time to go from state $n - \ell - 1$ to position $n - \ell + 1 < n$. By the~\cref{rem:phase-3}, the average time for the transitions $n - \ell - 1 \to n - \ell$ and $n - \ell \to n - \ell + 1$ to occur are $\O( n / \ell)$ for both, and from position $n - \ell$, unless \texttt{R} is chosen, it only takes $\O(n / (n - \ell)) = \O(1)$ time on average to fall back to state $n - \ell - 1$. Hence, the desired average time can be upper bounded by $\O(n^2 / \ell^2)$. Moreover, note that there is always a non-zero probability $p = \Omega(\ell^2 / n^2)$ to reach $n - \ell + 1$ from $n - \ell - 1$.

    Now, from position $n - \ell + 1$, if the event $E^3_t$ has already occurred then $\esp(T_3^1) = \O(n^2 / \ell^2)$ as desired. Otherwise, we define excursions starting in position $n - \ell + 1 < n$ and ending, when at some time $t$ either event $\texttt{H}_t^{n - \ell - 1}$ (that is, we fall back to state $n - \ell - 1$, which we consider as a failure) or $E^3_t$ (a success) occurs\footnote{And after a successful excursion, we stop tracking these excursions.}. First, as there is at most one successful excursion and a non-zero probability to reach $n - \ell + 1$ from $n - \ell - 1$ then from $n - \ell - 1$ we almost surely reach $n - \ell + 1$ in finite time, that is, a new excursion will almost surely occurs in finite time. From here, we can upper bound the average time between two excursions by the average time to go from state $n - \ell - 1$ to position $n - \ell + 1$, i.e., by $\O(n^2 / \ell^2)$ as we did earlier.

    Then we show that the number of failing excursions is at most $2$. To do so, observe that each excursion is preceded by a transition from $n - \ell$ to $n - \ell + 1$ and in each failing excursion, at least one step in the plateau is performed. Assume that at least $3$ failing excursions have occurred. This represents at least $3 \times 2 = 6$ transitions in the right plateau of \Jump, performed with \texttt{L} or \texttt{J} since for any failing excursion \RightBridge cannot be used to move forward in the plateau $[n - \ell .. n - 1]$ and it also rejects any move directed away of $x^*$, that is, toward $n - \ell - 1$. Moreover, by~\cref{lem:phase-3-}, at most two of these $6$ transitions can be performed with \texttt{L}. Thus, in one of the $3$ walks in the plateau $[n - \ell .. n - 1]$ (those corresponding to each of these failing excursions) at least the transitions $n - \ell \to n - \ell + 1$ and $n - \ell + 1 \to n - \ell$ should have been done with $\texttt{J}^+$ and $\texttt{J}^-$ respectively. As \texttt{J} as been used twice in this walk over the plateau $[n - \ell .. n - 1]$, it cannot be used anymore when coming back to $n - \ell$, that is, we cannot use \Jump to climb to state $n - \ell - 1$ as according to~\cref{lem:phase-3-}. From where, since \RightBridge does not accept this move, there is no way to reach $n - \ell - 1$ during this excursion. This is absurd hence, we cannot perform more than $2$ failing excursions.

    Also, \cref{lem:phase-3-} implies that objectives \texttt{L} and \texttt{J} can be used at most $4$ times (altogether) in a walk across the plateau $[n - \ell .. n - 1]$ and by~\cref{rem:phase-3} when accounting the average time to perform all transitions between neighboring positions, we deduce using \textsc{Wald}'s theorem~\cite{Wald44} and especially its simplified version from~\cite{DoerrK15} that a failing (resp. succeeding) excursion takes $\O(1)$ (resp. $\O(n / \ell)$) time on average hence $\esp(T_3^1) = \O(n^2 / \ell^2)$, that is, the runtime is mostly spent between the consecutive excursions.

    Now we derive the lower bound on $\esp(T_3^1)$. Consider the event
        \[ E = \{ T_1, T_2 < +\infty \} \cap \{ \| x_t \|_1 = n - \ell - 1, \, Q_t[0, \texttt{R}] = -\alpha r,\, Q_t[0, \texttt{J}] \geq 0 \}, \]
    where $t = T_1 + T_2 + 1$ which is finite in event $E$. As we proved in~\cref{thm:phase1} and~\cref{thm:phase2}, both $T_1$ and $T_2$ have finite expectation hence $\proba(T_1 < +\infty, T_2 < +\infty) = 1$ and
        \begin{align*}
            &\proba(E) \\
            &\  = \probacond{\| x_t \|_1 = n - \ell - 1, \, Q_t[0, \texttt{R}] = -\alpha r,\, Q_t[0, \texttt{J}] \geq 0}{T_1, T_2 < +\infty},
        \end{align*}
    where again $t = T_1 + T_2 + 1$. Then at time $T_1 + T_2$ we hit position $n - \ell$ for the first time and moreover $Q_{T_1 + T_2}[0, \texttt{L}] > 0$ while $Q_{T_1+T_2}[0, \texttt{J}]$ and $Q_{T_1 + T_2}[0, \texttt{R}]$ are still in $\{0, -\alpha r\}$ by~\cref{lem:NotMuchMistakes} and~\cref{lem:bounds} because entries $[0, \texttt{R}]$ and $[0, \texttt{J}]$ have never been updated during the second phase. Hence, \LeftBridge is selected\footnote{Another way to prove this fact is to invoke~\cref{lem-one-positive-per-state} because we know that $Q_{T_1 + T_2}[0, \texttt{L}] > 0$ hence, necessarily $Q_{T_1+T_2}[0, \texttt{J}]$ and $Q_{T_1 + T_2}[0, \texttt{R}]$ are non-positive.} at time $T_1 + T_2$, i.e., $f_{T_1 + T_2} = \texttt{L}$. Moreover, as we saw during the first phase, the only way to have both $Q_t[0, \texttt{R}] = -\alpha r$ and $Q_t[0, \texttt{J}] \geq 0$ is that event $\texttt{R}_0^+$ have occurred and objective \texttt{J} should have never been selected during the first phase (otherwise, we would have $Q_t[0, \texttt{J}] = -\alpha r$ since $[0..\ell]$ is a plateau for the \Jump function) hence, we can write
        \begin{align*}
            &\proba(E) \\
            &\ = \probacond{\| x_t \|_1 = n - \ell - 1, \, Q_t[0, \texttt{R}] = -\alpha r,\, Q_t[0, \texttt{J}] \geq 0}{T_1, T_2 < +\infty} \\ 
            &\ = \probacond{Q_t[0, \texttt{R}] = -\alpha r,\, Q_t[0, \texttt{J}] \geq 0}{T_1, T_2 < +\infty} \\
            &\qquad \times \probacond{\| x_t \|_1 = n - \ell - 1}{Q_t[0, \texttt{R}] = -\alpha r,\, Q_t[0, \texttt{J}] \geq 0, \, T_1, T_2 < +\infty} \\ 
            &\ \geq \left( \frac{1}{3} \cdot \frac{1}{2} \frac{n - 1}{n} \right) \cdot \frac{n - \ell}{n},
        \end{align*}
    where $\probacond{Q_t[0, \texttt{R}] = -\alpha r,\, Q_t[0, \texttt{J}] \geq 0}{T_1, T_2 < +\infty}$ has been lower bounded by the probability to use \RightBridge at time $t = 0$ and from position $1$ to use \LeftBridge and move toward position $2$ directly, which gives the factor $ \frac{1}{3} \cdot \frac{1}{2} \frac{n - 1}{n}$. For the other conditional probability, as we necessarily use \LeftBridge the first time we arrive in position $n - \ell$, whatever the value of the entries $[0, \texttt{R}]$ and $[0, \texttt{J}]$ then, we can get rid of the dependency on both $Q_t[0, \texttt{R}] = -\alpha r$ and $Q_t[0, \texttt{J}] \geq 0$ from where
        \begin{align*}
            &\probacond{\| x_t \|_1 = n - \ell - 1}{Q_t[0, \texttt{R}] = -\alpha r,\, Q_t[0, \texttt{J}] \geq 0, \, T_1, T_2 < +\infty} \\ 
            &\qquad = \probacond{\texttt{L}^-_{T_1 + T_2}}{T_1, T_2 < +\infty} \\ 
            &\qquad = \frac{n - \ell}{n} > \frac{1}{2},
        \end{align*}
    since $\ell < \frac{n}{2}$. Also, as $\frac{n-1}{n} > \frac{1}{2}$ we finally have the lower bound $\proba(E) \geq \frac{1}{3 \cdot 2 \cdot 2 \cdot 2} = \frac{1}{24} = \Omega(1)$, thus
        \[ \esp(T_3^1) \geq \proba(E) \espcond{T_3^1}{E} = \Omega\left( \espcond{T_3^1}{E} \right). \]

        Now, we need to lower bound $\espcond{T_3^1}{E}$ which, based on $E$, can be lower bounded by the average time to go from $n - \ell - 1$ to $n - \ell + 1$ knowing that $Q_t[0, \texttt{R}] = -\alpha r < 0 \leq Q_t[0, \texttt{J}]$ and $Q_t[0, \texttt{L}] < 0$ since \LeftBridge was used to move from $n - \ell$ to $n - \ell - 1$. In this setting, according to~\cref{lem:phase-3-ineq} we are never stuck in state $n - \ell - 1$ and there is a probability $\O(\ell / n)$ to move toward $n - \ell$ and, from this position, as \Jump will accept the move from either side, there is a probability $\frac{n - \ell}{n}$ to fall back to $n - \ell - 1$ (in which case, we keep using \Jump when we will hit $n - \ell$ again) and a probability $\frac{\ell}{n}$ to reach $n - \ell + 1$. Then, if we denote $\tau_{n - \ell - 1}$ (resp.  $\tau_{n - \ell}$ ) the average time to reach $n - \ell + 1$ from $n - \ell - 1$ (resp. $n - \ell$), we have
            \[ \tau_{n - \ell - 1} = 1 + \O(\ell / n)\tau_{n - \ell} + (1 - \O(\ell / n)) \tau_{n - \ell - 1}, \]
        that is $\tau_{n - \ell - 1} = \Omega(n / \ell) + \tau_{n - \ell}$, while 
            \[ \tau_{n - \ell} = 1 + \left( \frac{n - \ell}{n} \right) \tau_{n - \ell - 1} = 1 + \Omega((n - \ell) / \ell) + \left( \frac{n - \ell}{n} \right) \tau_{n - \ell}. \]
        Hence,
            \[ \tau_{n - \ell} = \frac{n}{\ell} + \Omega\left( \frac{n (n - \ell)}{\ell^2} \right) = \Omega(n^2 / \ell^2), \]
        since $\ell < \frac{n}{2}$ so $n - \ell \geq \frac{n}{2}$. Thus we obtain the lower bound
            \[ \esp(T_3^1) = \Omega(n^2 / \ell^2). \]
            \vspace*{-1cm}
\end{proof}

\end{document}


\title{Unlearning Works Better Than You Think: \\ Local Reinforcement-Based Selection of Auxiliary Objectives \\ {\large Supplementary materials: proofs and further experiments}
    \thanks{Identify applicable funding agency here. If none, delete this.}
    }
    
    \author{\IEEEauthorblockN{1\textsuperscript{st} Given Name Surname}
    \IEEEauthorblockA{\textit{dept. name of organization (of Aff.)} \\
    \textit{name of organization (of Aff.)}\\
    City, Country \\
    email address or ORCID}
    \and
    \IEEEauthorblockN{2\textsuperscript{nd} Given Name Surname}
    \IEEEauthorblockA{\textit{dept. name of organization (of Aff.)} \\
    \textit{name of organization (of Aff.)}\\
    City, Country \\
    email address or ORCID}
    \and
    \IEEEauthorblockN{3\textsuperscript{rd} Given Name Surname}
    \IEEEauthorblockA{\textit{dept. name of organization (of Aff.)} \\
    \textit{name of organization (of Aff.)}\\
    City, Country \\
    email address or ORCID}
    }
    
    \maketitle
    
    \begin{abstract}
        This is 
    \end{abstract}
    
    \begin{IEEEkeywords}
        evolutionary algorithm, single-objective optimization, reinforcement learning
    \end{IEEEkeywords}